\documentclass[acmtog, screen, nonacm
]{templates/acmart/acmart}

\newcommand{\status}{0}

\input{conf/packages}
\definecolor{INCOMPLETECOLOR}{RGB}{178,34,34}
\definecolor{UNDERREVISIONCOLOR}{RGB}{210,121,121}
\definecolor{FEEDBACKNEEDEDCOLOR}{RGB}{230,170,50}
\definecolor{FEEDBACKGIVENCOLOR}{RGB}{121,210,121}
\definecolor{COMPLETECOLOR}{RGB}{121,124,210}
\definecolor{LOCKEDCOLOR}{RGB}{153,102,255}

\definecolor{TODOCOLOR}{RGB}{255,0,0}
\definecolor{MONDECOLOR}{RGB}{0,0,255}
\definecolor{QICOLOR}{RGB}{118,185,0}
\definecolor{ANJULCOLOR}{RGB}{127,127,0}
\definecolor{RACHELCOLOR}{RGB}{127,0,127}
\definecolor{GUESTCOLOR}{RGB}{0,127,127}
\definecolor{YUJIECOLOR}{RGB}{118,185,0}
\definecolor{WHITE}{RGB}{255,255,255}

\newcommand{\nothing}[1]{}
\newcommand{\isolated}[1]{\hfill\break#1\xspace}

\ifthenelse{\equal{\status}{0}} {                     

    \newcommand{\note}[1]{{\it\color{blue} #1}}

    \newcommandx{\monde}[2][1=]
        {\setulcolor{MONDECOLOR}{\ul{#1}}
         \isolated{\textcolor{MONDECOLOR}{\textbf{Monde:} #2}}}
    \newcommandx{\qisun}[2][1=]
        {\setulcolor{QICOLOR}{\ul{#1}}
         \isolated{\textcolor{QICOLOR}{\textbf{Qi:} #2}}}
    \newcommandx{\anjul}[2][1=]
        {\setulcolor{ANJULCOLOR}{\ul{#1}}
         \isolated{\textcolor{ANJULCOLOR}{\textbf{Anjul:} #2}}}
    \newcommandx{\rachel}[2][1=]
        {\setulcolor{RACHELCOLOR}{\ul{#1}}
         \isolated{\textcolor{RACHELCOLOR}{\textbf{Rachel:} #2}}}

    \newcommandx{\guest}[3][1=]
        {\setulcolor{LOCKEDCOLOR}{\ul{#1}} \textcolor{LOCKEDCOLOR}
        {[\textbf{#2:} #3]}}
}{                                                    
    \newcommand{\warning}[1]{}
    \newcommand{\note}[1]{}
    
    \newcommandx{\monde}[2][1=]{#1}
    \newcommandx{\qisun}[2][1=]{#1}
    \newcommandx{\anjul}[2][1=]{#1}
    \newcommandx{\rachel}[2][1=]{#1}
    \newcommandx{\guest}[3][1=]{#1}
         \newcommandx{\yujie}[2][1=]
        {\setulcolor{YUJIECOLOR}{\ul{#1}}
    }
    }

\ifthenelse{\equal{\status}{0}} {          
    
}{}
\ifthenelse{\equal{\status}{1}} {          
    
}{}
\ifthenelse{\equal{\status}{2}} {          
    
}{}
\ifthenelse{\equal{\status}{3}} {          
    
}{}
\ifthenelse{\equal{\status}{4}} {          
    
}{}

\newcommand{\greencheck}{{\color{Green}\cmark}\xspace}
\newcommand{\green}{\cellcolor{Green!12.5}\greencheck}
\newcommand{\yellowcheck}{{\color{YellowOrange}(\cmark)}\xspace}
\newcommand{\yellow}{\cellcolor{YellowOrange!12.5}\yellowcheck}
\newcommand{\redcheck}{{\color{red}\xmark}\xspace}
\newcommand{\red}{\cellcolor{red!12.5}\redcheck}

\newcommand{\pc}[1]{\textcolor{blue}{#1}}
\definecolor{bestgray}{gray}{0.65}   
\definecolor{secondgray}{gray}{0.80} 
\definecolor{thirdgray}{gray}{0.92}  

\setcopyright{none}
\renewcommand{\footnotetextcopyrightpermission}[1]{}
\title{DeltaCam: Differential Intrinsic Camera Modeling for Video Generation}

\author{Debabrata Mandal}
\email{debman@cs.unc.edu}
\affiliation{%
  \institution{UNC, Chapel Hill}
  \country{USA}
}

\author{Zhihan Peng}
\affiliation{%
  \institution{UNC, Chapel Hill}
  \country{USA}}
\email{zp59@unc.edu}

\author{Yujie Wang}
\affiliation{%
  \institution{UNC, Chapel Hill}
  \country{USA}
}

\author{Praneeth Chakravarthula}
\affiliation{%
 \institution{UNC, Chapel Hill}
 \country{USA}}

\renewcommand{\shortauthors}{Mandal et al.}


\begin{document}
\begin{teaserfigure}
  \includegraphics[width=\textwidth]{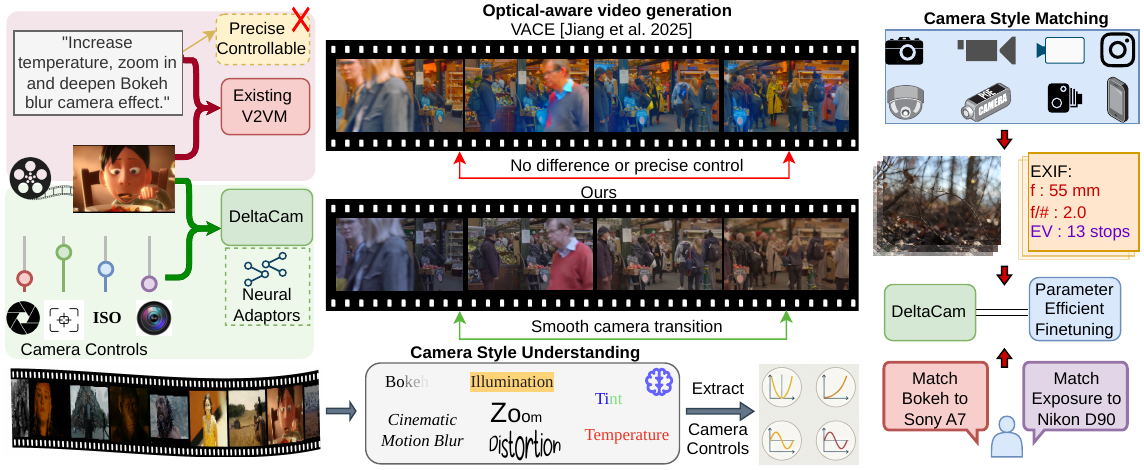}
  \caption{\textit{Teaser figure.} We present a fully integrated camera controlled video generation model including photographic and cinematographic effects such as \textit{Bokeh} and Dolly zoom, while preserving original scene dynamics. Our method proposes a novel architectural block to disentangle camera conditioning for different parameters jointly during inference. Further, we also propose style extraction from videos for photographic concepts enabling video-to-video style transfer. Finally, we demonstrate our method can easily adapt its imaging process to diverse real camera types from just paired image camera EXIF metadata.}
  \Description{}
  \label{fig:teaser}
\end{teaserfigure}

\begin{abstract}

Incorporating camera intrinsics into video generation models offers a principled way to control not only scene dynamics but also the imaging process that governs visual appearance.
Prior work has primarily focused on extrinsic control, such as camera pose and motion, while treating intrinsic camera parameters as implicit or fixed.
A key bottleneck is the lack of large-scale video datasets with accurate and diverse temporally varying camera metadata, which makes learning absolute camera parameterizations difficult.
As a result, current models struggle to incorporate photographic camera behavior, including depth-of-field transitions, exposure variations, lens distortions, and color processing, in a controllable and temporally consistent manner.


We introduce DeltaCam, a video diffusion framework that models camera behavior through $\Delta$-parameterized neural camera adaptors, operating on relative changes in camera motion and intrinsics instead of absolute states. By learning this differential formulation from synthetic video data, we mitigate reliance on precise real-world camera labels and enable smooth, consistent control over imaging factors such as focal length, aperture, ISO, color temperature, and lens distortion. We extend this framework to real-world footage through two mechanisms: finetuning the controls on real image-metadata pairs for precise shot matching, and extracting disentangled embeddings for implicit video-to-video style transfer without requiring explicit camera parameters. By effectively separating scene content from intrinsic imaging behavior, DeltaCam enables camera-consistent video generation and editing operations that are difficult to achieve with existing models. Ultimately, our results establish a practical and scalable approach for bridging synthetic control and real-world photographic emulation.
\end{abstract}

\maketitle


\section{Introduction}
\label{sec:introduction}

\begin{figure*}[t]
    \includegraphics[width=\textwidth]{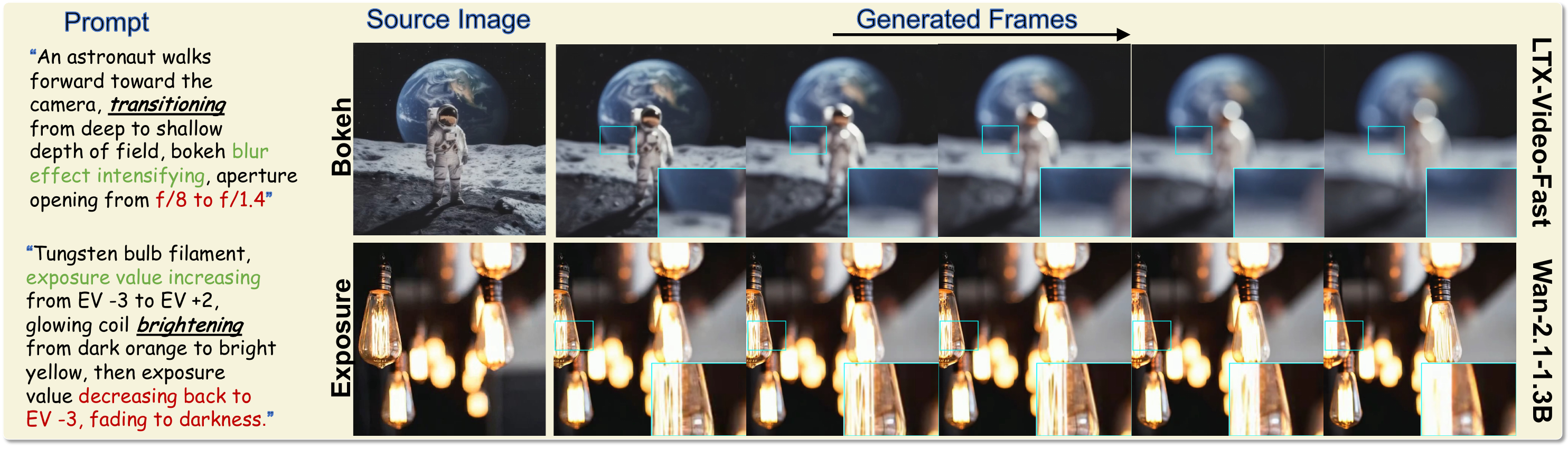}
    \caption{\textit{Limitations of current models.} We investigate the performance of prior text-to-video models \cite{HaCohen2024LTXVideoRV, Wang2025WanOA} with camera heavy text prompts to achieve accurate camera transitions. Text in \textcolor{YUJIECOLOR}{green} indicates prompt followed, while text in \textcolor{red}{red} represents prompt not followed.}
    \label{fig:pilot}
\end{figure*}

Modern video generation models lack a principled and editable representation of camera behavior, a core component of real-world image formation and videography. 
In existing approaches \cite{bahmani2025ac3d, zhou2025stable, wang2025akira}, scene dynamics, camera motion, and camera intrinsics are all entangled within a single implicit latent space.
While sufficient for casual video synthesis, this entanglement severely limits controllability and interpretability of camera behavior. 
In practice, this prevents fundamental operations like maintaining a consistent lens curvature across a generated shot or smoothly transitioning a camera parameter over time; the capabilities essential for real-world video creation and post-production editing. 
To verify this, we ran a simple experiment shown in \cref{fig:pilot} where we use text prompts, crafted with photography keywords to guide the video model generation. We find most state-of-the-art models fail to follow the user prompts accurately and sometimes even grossly deviating from the provided instructions.
This limitation mainly stems from severe data constraints: large-scale video datasets rarely provide accurate, temporally consistent camera metadata, and camera intrinsics are often left fixed, unreliable, or not recorded at all. As a result, learning absolute camera parameterization is not only challenging but also difficult to scale.
Although most recent works have explored synthetic camera pipelines for controllable image generation \cite{fang2024camera,yuan2025generative}, these approaches do not extend naturally to temporally coherent video. 
Moreover, they often suffer from camera-parameter--image-appearance mismatches introduced by synthetic-to-real domain gap. 

Here, we introduce \textit{DeltaCam}, a video diffusion framework that models camera behavior through $\Delta$-parameterized differential neural camera adaptors.
Our framework is motivated by a key observation that cameras are \textit{not} operated through absolute states, but through \textit{relative changes}: adjusting zoom, aperture, exposure, or color settings gradually. We achieve this by varying a single camera parameter at a time during the video capture process in a smooth and continuous manner to mimic real world camera footage. Since the video backbone model's core layers remain frozen, it implicitly learns to disentangle the photographic (camera intrinsic) concepts in the latent space. To better understand and control photographic concepts in video generation, we propose a disentangled camera control architecture where camera dynamics and scene information are controlled along different axes unlike prior works which do them jointly through spatial conditioning \cite{yuan2025generative}. Furthermore, we extract temporally-varying camera style embeddings from videos disentangled from the scene content unlike prior works \cite{radford2021learning, oquab2023dinov2}. 



Finally, we apply our camera style understanding and camera controlled
video generation framework to applications such as video-to-video style transfer for photographic effects and camera style matching. Our contributions are summarized below,
\begin{itemize}[noitemsep,topsep=0pt,parsep=0pt,partopsep=0pt,leftmargin=*]
    \item We \textit{introduce} DeltaCam, a generative video framework that models camera behavior through $\Delta-$parameterized differential neural adaptors, representing relative changes in camera motion and intrinsics.
    \item We \textit{develop} a disentangled camera representation that separates camera extrinsics from intrinsics, while remaining robust to sparse and heterogeneous camera supervision.    
    \item We \textit{demonstrate} camera-consistent video synthesis and editing, enabling temporally smooth and transferable camera behavior that existing video generation models struggle to support.
    \item We further showcase several downstream applications using our model, including video-to-video style transfer and \textit{parameter-efficient} camera style matching scalable to diverse new cameras.
\end{itemize}

\section{Related Work}
\label{sec:related}

As summarized in Table~\ref{tab:related_work}, prior work in generative camera control and optical modeling primarily focuses on geometric motion or image-level edits, leaving intrinsic camera behavior for video generation largely unaddressed.

\subsection{Camera-Controlled Video Generation}

Recent video generation works incorporate camera control primarily via extrinsics like viewpoint, pose, and trajectory \cite{bahmani2025ac3d, li2025realcam, zhou2025stable,bai2025recammaster}. 
Training-based approaches inject explicit camera or geometry representations into diffusion models \citep{blattmann2023stable, yang2024cogvideox}. These include ray-based embeddings (e.g., Pl\"ucker coordinates in AC3D \cite{bahmani2025ac3d} and SeVA \cite{zhou2025stable}) and point-cloud renderings \cite{yu2024viewcrafter, xing2025motioncanvas}. ReCamMaster \citep{bai2025recammaster} trains a DiT on synthetic 4D data for cinematic transitions and refocusing.
Conversely, training-free methods adapt pretrained models to novel trajectories. CamTrol \citep{hou2024training} and LatentReframe \citep{zhou2025latent} use point-cloud reprojection as geometric guidance for inpainting, while CamMiMic \cite{guhan2025cammimic} transfers motion via homography-guided refinement without 3D reconstruction.

However, these methods treat camera control merely as geometric motion. Intrinsic imaging factors (focal length, aperture, ISO, color temperature, distortion) are absorbed into the video prior, limiting coherent synthesis of depth-of-field transitions, exposure ramps, white-balance shifts, and dolly zooms. DeltaCam addresses this by representing camera behavior as a temporally evolving control signal over both motion and intrinsic imaging parameters.

\begin{table}[!t]
    \setlength{\tabcolsep}{0em}
    \centering
    \footnotesize
    \caption{
        Comparison of related work on camera-controlled generation. Each criterion is fully~\greencheck, partially~\yellowcheck, or not met ~\redcheck. General Video-to-Video (V2V) editing models like VACE lack explicit, disentangled camera controls. 
    }
    \begin{tabularx}{\linewidth}{m{0.24\linewidth}XXXXXX}
    \toprule
    &
    {\footnotesize GenPhoto \shortcite{yuan2025generative}} &
    {\footnotesize AC3D ~\shortcite{bahmani2025ac3d}} &
    {\footnotesize Akira ~\shortcite{wang2025akira}} &
    {\footnotesize VACE \shortcite{jiang2025vace}} &
    {\footnotesize ReCamMaster \shortcite{bai2025recammaster}} &
    {\footnotesize \textbf{Ours}} \\
    \midrule
    \multicolumn{7}{l}{\textbf{Model Characteristics}}\\ 
    \midrule
    Video-to-Video & 
    \red & 
    \red & 
    \red & 
    \green & 
    \yellow &
    \green \\
    Extrinsic Control  & 
    \red & 
    \green & 
    \green & 
    \red & 
    \green &
    \green \\
    Intrinsic Control & 
    \green & 
    \red & 
    \yellow & 
    \red & 
    \red &
    \green \\
    Precise Control & 
    \yellow & 
    \red & 
    \yellow & 
    \red & 
    \yellow &
    \green \\
    Smooth Control & 
    \red & 
    \yellow & 
    \green &
    \yellow & 
    \green  &
    \green \\
    Style Mixing & 
    \yellow & 
    \red & 
    \red & 
    \red & 
    \red & 
    \green \\
    Style Transfer & 
    \red & 
    \red & 
    \red & 
    \yellow & 
    \red & 
    \green \\
    Camera Matching  & 
    \red & 
    \red & 
    \red &
    \red & 
    \red  &
    \green \\
    \bottomrule
\end{tabularx}
    \label{tab:related_work}
\end{table}

\begin{figure*}[!h]
\includegraphics[width=\textwidth]{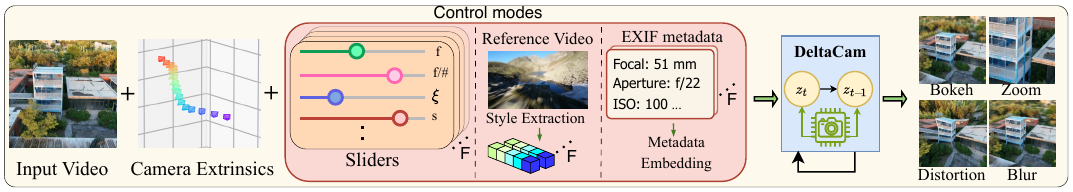}
    \caption{\textit{Method Overview}. We present a video-to-video generation pipeline with novel camera extrinsics and intrinsics controllable by \textit{per-frame} sliders, a reference video with \textit{photographic} styles, or real camera \textit{paired} image-EXIF metadata.}
    \label{fig:overview}
\end{figure*}

\subsection{Camera-Intrinsic Control in Generative Models}
Related works explore camera parameters and optical effects in generative image models. Early methods like NeuralCam~\citep{ouyang2021neural} treat optical effects as image-to-image translation, learning separate modules for aperture, ISO, and noise. Recent text-to-image models introduce camera-aware conditioning: Generative Photography~\citep{yuan2025generative} and CamTokens~\cite{fang2024camera} control focal length, aperture, and ISO, while PreciseCam~\cite{bernal2025precisecam} adds spatial control via ControlNet~\cite{zhang2023adding}. For video generation, AkiRa~\cite{wang2025akira} augments the standard pinhole camera model with aperture and lens-distortion modeling.

A parallel challenge is \emph{representing} camera parameters to generalize across devices, sensors, and lenses. CamTokens~\cite{fang2024camera} explores token embeddings for this; CCMNet~\cite{kim2025ccmnet} extracts camera-specific fingerprints for color constancy; and unified camera models aid perception tasks like depth estimation across diverse camera types~\cite{guo2025depth}. Additionally, Puffin~\cite{liao2025thinking} introduces a camera-centric latent space for prompt-based scene guidance, albeit without explicit optical or photographic modeling.

While these methods establish camera parameters as generative controls, they typically map \emph{absolute} camera states to visual effects. 
This requires precisely calibrated metadata—usually restricted to synthetic pipelines or curated still-image datasets—and ties mappings to specific sensor-lens combinations, limiting transfer to real video. Instead, DeltaCam maps \emph{relative} camera changes to visual changes. This inductive bias is roughly device-invariant, enables supervision from sparse, heterogeneous real-world video, and naturally supports transferring camera trajectories between videos.

\subsection{Cinematic Control and Video-to-Video Transfer}
Fine-grained video generation control relies heavily on structured spatial and motion signals, including 3D bounding boxes~\cite{Wang2025CineMasterA3}, 2D poses~\cite{xing2025motioncanvas}, edge maps~\cite{zhang2023controlvideo}, motion cues~\cite{Chen2023MotionConditionedDM}, and point tracks~\cite{Geng2024MotionPC}, with extensions to long-form narratives~\cite{xiao2025captain} and motion recovery from blur~\cite{tedla2025generating}. While effective for scene layout and object motion, these handles leave camera-dependent imaging implicit: the temporal modulation of focus, exposure, or focal length---core cinematic behaviors---cannot be specified via spatial control maps.

Complementary works target \emph{video-to-video transfer}, extracting reference attributes to re-apply to a target. SDEdit~\cite{meng2021sdedit} established training-free guided synthesis; VideoSwap~\cite{gu2024videoswap} transfers subjects via 2D-keypoints, DDIM inversion~\cite{song2020denoising}, and latent blending; DreamVideo~\cite{wei2024dreamvideo} fuses separately learned subject and motion adapters; and Diffusion-as-Shader~\cite{gu2025diffusion} transfers motion using 3D point tracks. Recently, DisMo~\cite{resslerdismo} extracted reusable motion representations disentangled from subject, appearance, and pose. While these methods successfully disentangle and transfer scene-level factors, none isolate the camera's intrinsic imaging behavior.

Whereas prior works transfer object motion, identity, or style, we target this unexplored axis of camera behavior. Through reusable $\Delta$-adapters, DeltaCam natively supports both a slider interface for direct parameter control and a reference-driven mode to extract and apply $\Delta$-trajectories across videos for shot matching and camera-style transfer.

\section{Method}
\label{sec:method}

This section details DeltaCam, our framework for integrating explicit, $\Delta$-parameterized intrinsic camera control into video generation. We first introduce the core camera formulation and Camera Conditioning Module (CCM) in \cref{sec:overview,sec:cond_video}, followed by our reference-style extraction and EXIF metadata matching techniques in \cref{sec:style_understanding,sec:matching}. The overall architecture is illustrated in \cref{fig:overview}, with the detailed CCM and multi-stage training curriculum (\cref{sec:training}) depicted in \cref{fig:pipeline,fig:camera_fusion_architecture}.




\subsection{Overview} 
\label{sec:overview}

We integrate a comprehensive, general (non-pinhole) camera model into the video generation pipeline, operable via three main control modes: \textit{parameter sliders}, \textit{style embeddings}, and \textit{EXIF metadata embeddings} (\cref{fig:overview}). Our primary focus is video-to-video generation guided by camera-centric controls that explicitly decouple camera trajectory (extrinsics) from photographic effects (intrinsics). 

Unlike standard pinhole models typically assumed in rendering pipelines, real cameras introduce complex optical and sensory transformations. We model this capture process as a function $\Phi$ governed by the underlying 3D scene $S$, extrinsic pose $\mathbf{P} \in SE(3)$, and a $K$-dimensional vector of intrinsic photographic parameters $\boldsymbol{\theta} \in \mathbb{R}^K$ (e.g., focal length, aperture, ISO):
\begin{equation}
    I = \Phi(S, \mathbf{P}, \boldsymbol{\theta})
    \label{eq:cam}
\end{equation}

Inverting \cref{eq:cam} to estimate these intrinsic parameters from raw video is highly ill-posed. However, solving this inverse problem is essential for enabling video-to-video camera style transfer when explicit metadata is unavailable. 

The following subsections detail our approach for conditioning the video generative model for full camera-aware synthesis, extracting these camera styles directly from reference footage, and validating the model's generalization by matching the shot profile of real-world cameras using paired image-EXIF data.

\begin{figure}[h]
\includegraphics[width=\columnwidth]{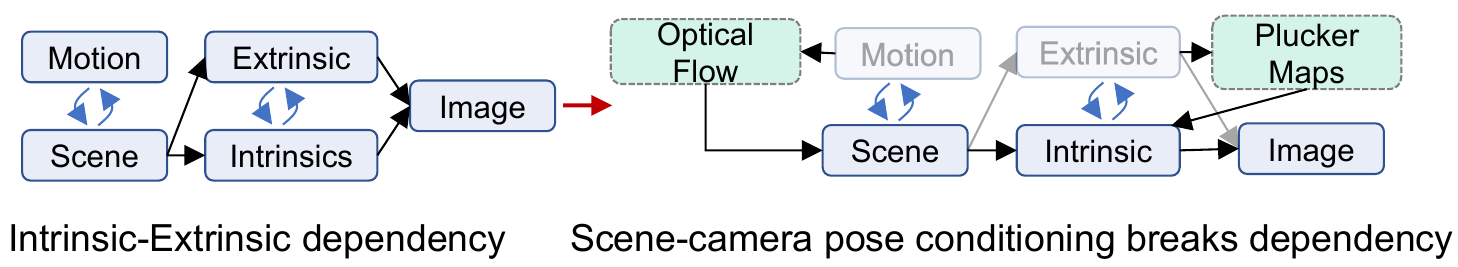}
    \caption{\textit{Camera conditioning circular dependency}. Real videos are captured in dynamic environments with moving objects or moving camera leading to a criss-cross dependency in the final imaging process. For video generation, we break this circular dependency by conditioning the model on optical flow and per-frame camera extrinsic for generating the video along a different camera trajectory.}
    \label{fig:motivation}
\end{figure}
\vspace{-0.5cm}
\subsection{Camera video conditioning}  
\label{sec:cond_video}
\paragraph{Architecture Motivation.} 

The camera capture process in \cref{eq:cam} mixes any change in scene, extrinsics, or intrinsics in any order (\cref{fig:motivation}). We break this entanglement by anchoring scene content via the source video and conditioning generation on geometric proxies
%
$\mathbf{G}$ (depth, optical flow, and perspective fields) to serve as the spatial backbone, helping the model identify structure, motion, and orientation. To ensure consistent viewpoint rendering, we encode the new extrinsic camera pose (relative to starting frame) $\mathbf{P}$ as spatial Pl\"ucker ray maps \cite{sitzmann2021light} and inject them into both the intrinsic conditioning branch and the base video diffusion model. We zero out the pose ray maps in the case of same video camera style editing (no novel view) and keep this behavior consistent during inference. The proxy maps $\mathbf{G}$ provide additional 3D scene cues over the RGB video.

\begin{figure}[h]
\includegraphics[width=\columnwidth]{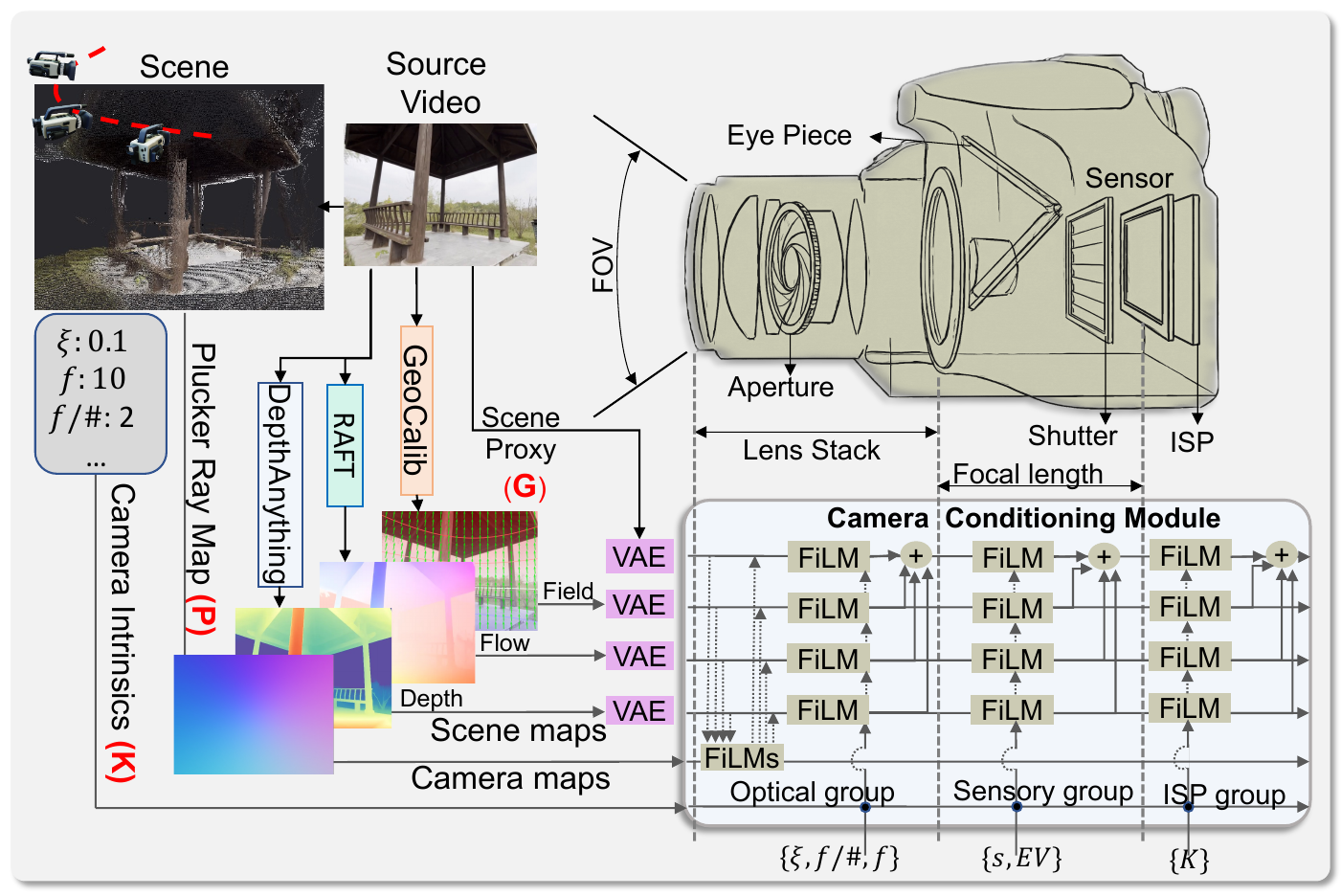}
    \caption{\textit{Camera Conditioning Module}. Our proxy camera model works by disentangling the optical, sensory and image characteristics (K) of regular hand held cameras. Scene maps (G) such as depth, optical flow and perspective fields are used to guide the video generation process from our camera model as a substitute for the real 3D world.}
    \label{fig:pipeline}
    \vspace{-0.5cm}
\end{figure}

Accordingly, we design our Camera Conditioning Module (CCM) to modulate these 3D spatial cues based on user-defined photographic trajectories (\cref{fig:pipeline}). However, absolute camera parameters are ambiguous without knowing the camera type, which alters how each parameter maps to the final appearance. We therefore adopt a $\Delta$-parameterization: given an intrinsic trajectory, we define the relative control at time $t$ with respect to the anchor frame as $\Delta\boldsymbol{\theta}_t = (\boldsymbol{\theta}_t - \boldsymbol{\theta}_1) / r$, where $r$ is the parameter's maximum range. This (a) decouples training from the specific camera, since a $\Delta$ directly maps to a change in effect, and (b) cleanly isolates individual effects (e.g., Bokeh, zooming) by varying one $\Delta\theta_i$ while holding the rest at zero (i.e., locked to the anchor frame).

The spatial proxy maps $\mathbf{G}_t$ are first compressed into latent features $\mathbf{h}_t$ via a frozen 3D VAE encoder. We then modulate these spatial features using Feature-wise Linear Modulation (FiLM) \cite{Perez2017FiLMVR} driven by the relative intrinsics:
\begin{equation}
\text{FiLM}(\mathbf{h}_t, \Delta\boldsymbol{\theta}_t) = \gamma(\Delta\boldsymbol{\theta}_t) \odot \mathbf{h}_t + \beta(\Delta\boldsymbol{\theta}_t),
\end{equation}
where $\gamma$ and $\beta$ are scaling and shifting tensors predicted by a multi-layer perceptron (MLP) from the relative control input $\Delta\boldsymbol{\theta}_t$. These FiLM layers use simple affine transformations to adjust the optical style globally. Because they preserve the underlying scene structure, they naturally encourage the model to disentangle camera effects from scene content in the latent space. 
We split the intrinsics into optical, sensory, and ISP groups, each routed through a separate FiLM cascade (\cref{fig:pipeline}). Cascading these FiLM layers across the encoded spatial feature maps aggregates the final intrinsic conditioning before fusing it with the base video latent representation $\mathbf{z}_t$.

\subsection{Reference Camera Style Extraction} 
\label{sec:style_understanding}
\begin{figure}[h]
\includegraphics[width=\columnwidth]{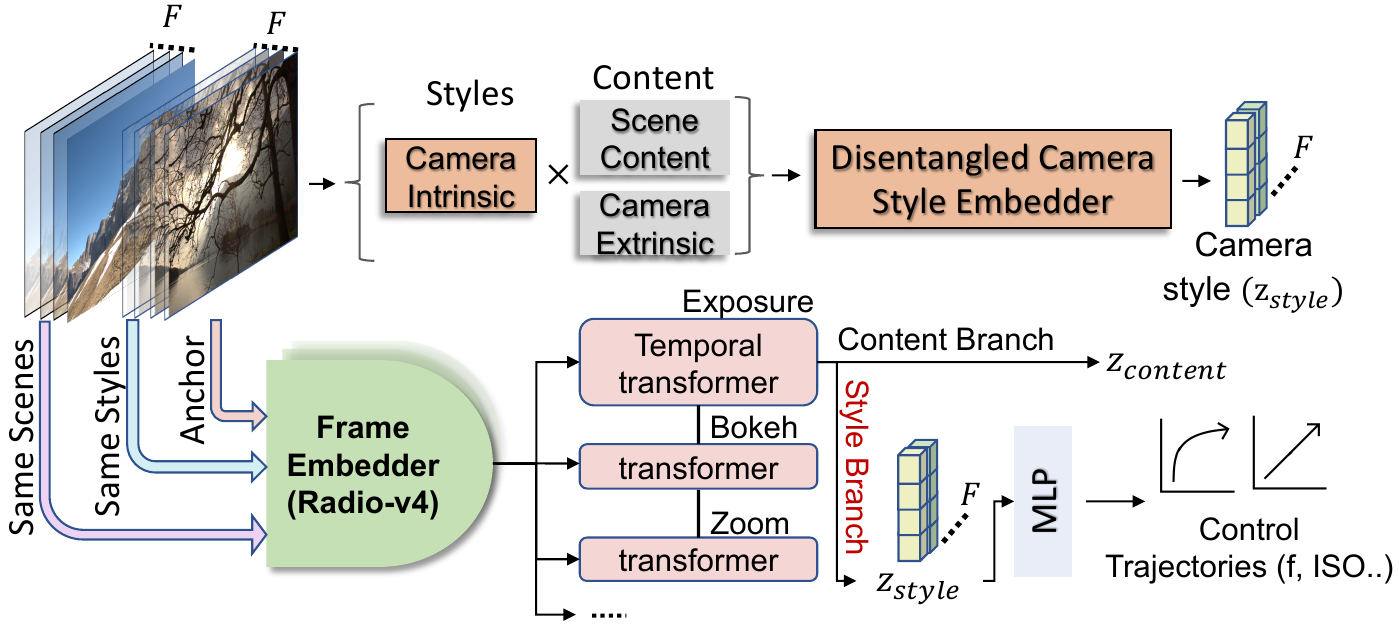}
    \caption{\textit{Camera Style Extraction.} Our camera style embedder extracts styles from video by differentiating between the content and styles from a batch of videos. The styles are extracted on a per-frame basis for each common photographic effects (e.g. Bokeh, zooming) using a shared temporal transformer. The final styles are projected into 1D camera parameters (e.g. f, aperture) by linear MLP layers. }
    \label{fig:style_extraction}
\end{figure}

We extract reusable camera style embeddings from videos by explicitly disentangling spatial representations into orthogonal content and style branches (\cref{fig:style_extraction}). To supervise this disentanglement, we construct training batches comprised of synthetic video triplets $(\mathbf{x}_a, \mathbf{x}_c, \mathbf{x}_s)$ generated via our synthetic camera pipeline. This synthetic environment is crucial, as it allows us to perfectly isolate variables that are otherwise impossible to decouple in real-world footage. Specifically, the anchor video $\mathbf{x}_a$ shares the exact scene content with $\mathbf{x}_c$ but under a different camera trajectory, while $\mathbf{x}_s$ shares the exact camera trajectory with $\mathbf{x}_a$ but depicts an entirely different scene:
\begin{equation}
  \mathbf{x}_a \xrightarrow{\text{same scene}}   \mathbf{x}_c, \qquad
  \mathbf{x}_a \xrightarrow{\text{same style}}   \mathbf{x}_s.
\end{equation}

We extract spatial patch features from each clip with a frozen c-RADIO ViT-H backbone\cite{ranzinger2026cradiov4techreport}, and subsequently aggregate them using a lightweight temporal transformer. The network then bifurcates to decode a scene content embedding $\mathbf{z}_c$ and a per-frame style embedding $\mathbf{z}_s$. A dedicated regression head maps $\mathbf{z}_s$ to the predicted intrinsic trajectory $\hat{\boldsymbol{\tau}}$ (representing the temporal evolution of our camera parameters).


This structure drives the disentanglement objective: the content branch is trained to maximize the similarity between the content embeddings of $(\mathbf{x}_a, \mathbf{x}_c)$, while the style branch aligns the style embeddings of $(\mathbf{x}_a, \mathbf{x}_s)$. The overall style extraction loss function is:
\begin{equation}
  \mathcal{L} =
    \lambda_\tau \mathcal{L}_\mathrm{NCC}(\hat{\boldsymbol{\tau}},\boldsymbol{\tau})
  + \lambda_c   \mathcal{L}_\mathrm{NCE}(\mathbf{z}_c^a, \mathbf{z}_c^c)
  + \lambda_s   \mathcal{L}_\mathrm{NCE}(\mathbf{z}_s^a, \mathbf{z}_s^s)
  + \lambda_\mathrm{MI} \mathcal{L}_\mathrm{MI}(\mathbf{z}_c,\mathbf{z}_s)
\end{equation}
where $\mathcal{L}_\mathrm{NCC}$ measures trajectory prediction accuracy via Normalized Cross-Correlation against the ground truth trajectory $\boldsymbol{\tau}$, and $\mathcal{L}_\mathrm{NCE}$ is the standard InfoNCE contrastive loss term \cite{radford2021learning}. To explicitly penalize residual dependence between the content and style branches, $\mathcal{L}_\mathrm{MI}$ minimizes the squared cosine similarity between content and style embeddings.
%
To populate this training data, we leverage our synthetic camera pipeline to sample diverse trajectories across commonly used photographic effects, including \textit{Bokeh}, dolly zooms, lens curvature, exposure ramps, cinematic blur, and color temperature.

\vspace{-0.5cm}

\subsection{Camera Style Matching} 
\label{sec:matching}

While DeltaCam relies on relative $\Delta$-parameterization learned from synthetic data, practical applications often require grounding the model to absolute, real-world camera states. We achieve this through \textit{Camera Style Matching}, which adapts our synthetic-trained controls using real image-metadata pairs for precise, camera-specific shot emulation.

Given a vector of absolute per-frame parameters $\mathbf{m}_t \in \mathbb{R}^{N_{\text{meta}}}$ derived from image EXIF metadata (e.g., aperture $f$, focal length $\ell$, ISO speed $s$), we first map each physical quantity to a normalized scalar to reflect perceptual linearity (where equal steps correspond to equal perceptual increments, such as one f-stop or one EV). We define the normalized vector $\tilde{\mathbf{m}}_t \in [0,1]^{N_{\text{meta}}}$ via a log-space mapping for each element $i$:
\begin{equation}
    \tilde{m}_{t,i} = \frac{\log(m_{t,i} / m_{\min,i})}{\log(m_{\max,i} / m_{\min,i})}
\end{equation}

A lightweight EXIF metadata tokenizer $\Phi_{\text{meta}}: \mathbb{R}^{N_{\text{meta}}} \to \mathbb{R}^{d_\xi}$ implemented as a two-layer MLP with a SiLU activation and a zero-initialized output layer—maps this normalized state to a latent style embedding $\boldsymbol{\xi}_t = \Phi_{\text{meta}}(\tilde{\mathbf{m}}_t)$. This generates a continuous per-clip metadata sequence $\boldsymbol{\xi} \in \mathbb{R}^{T \times d_\xi}$ that integrates directly into our conditioning pipeline.

\begin{figure}[h]
\includegraphics[width=\columnwidth]{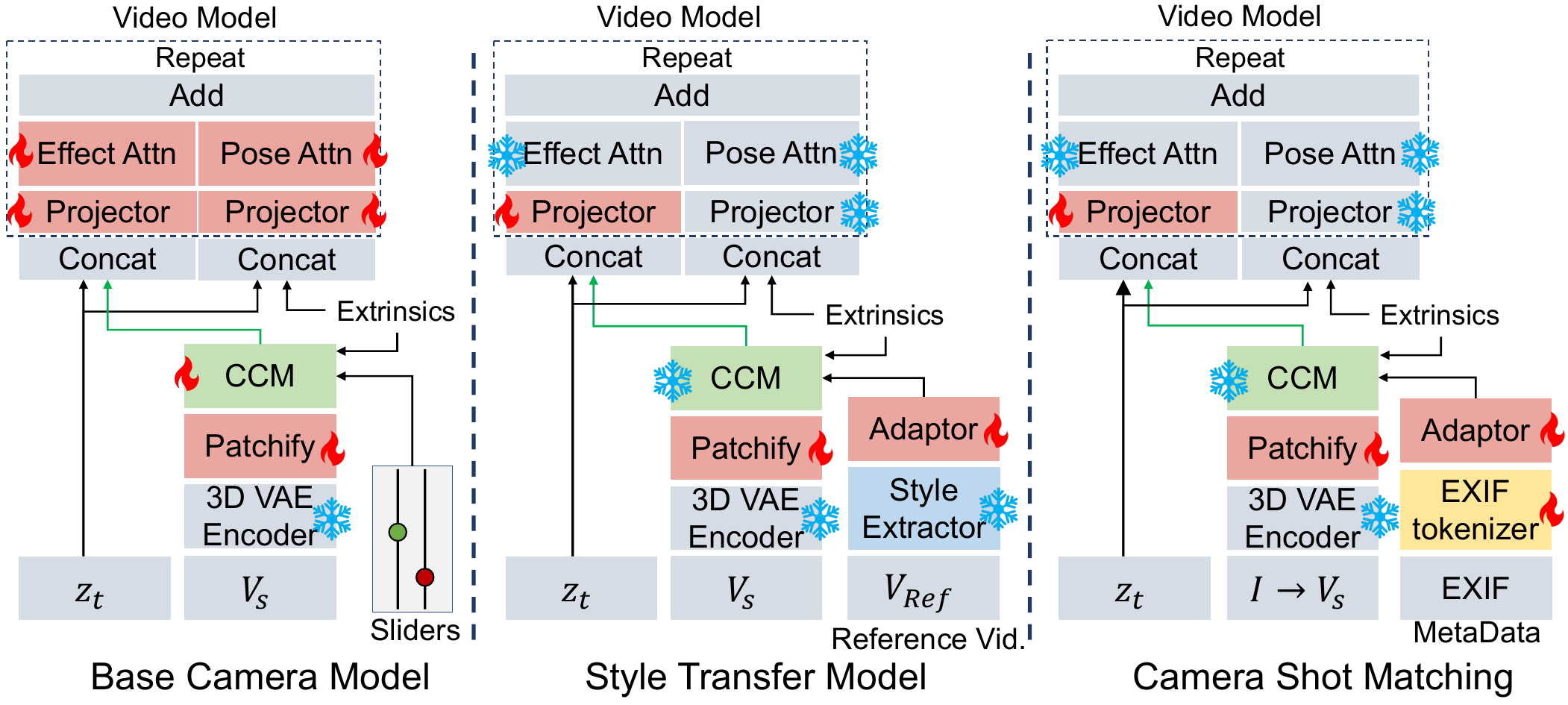}
    \caption{\textit{Camera-VDM integration.} We condition the VDM on camera extrinsics and intrinsics (through CCM) separately from one another. Style transfer and camera style matching are achieved through lightweight adaptors added prior to the camera encoder (CCM).}
    \label{fig:camera_fusion_architecture}
    \vspace{-0.5cm}
\end{figure}


\subsection{Training Curriculum} 
\label{sec:training}

We employ a three-stage curriculum to align the base Video Diffusion Model (VDM) with our camera-aware framework without degrading its innate generative priors (\cref{fig:camera_fusion_architecture}):

\begin{enumerate}[itemsep=0pt, leftmargin=*]
    \item \textit{Base pre-training:} We train the Camera Conditioning Module (CCM) and a set of new, independent self-attention blocks using synthetic data. To preserve the base VDM's high-quality video generation, we leave all of its original layers completely frozen. The model processes the extrinsic pose $\mathbf{P}$ and relative intrinsic trajectories $\Delta\boldsymbol{\theta}_t$ by concatenating them with the video latent $\mathbf{z}_t$ and passing them exclusively through the new attention blocks.
    
    \item \textit{Style transfer adaptation:} We freeze the VDM, the core CCM, and the new attention blocks. We then train only lightweight style extractor, teaching it to map reference embeddings extracted trajectory into the prior learnt $\Delta$-parameter control space.
    
    \item \textit{Real-world matching:} We keep the entire network (from stage 1) frozen and train only the EXIF metadata tokenizer $\Phi_{\text{meta}}$. This step translates the absolute normalized sequence $\boldsymbol{\xi}_t$ into our synthetic control space, allowing the model to accurately emulate real cameras. Real world camera adaptation requires updating less than 0.1\% of the overall model parameters allowing \textit{parameter efficient finetuning} of the diffusion model to diverse camera types.
\end{enumerate}

Extended architectural details are provided in the supplementary.




\section{Experiments}
\label{sec:result}


\subsection{Dataset \& Implementation}

 \label{sec:train} 

We train on a 50:50 mix of RealCamVid~\cite{li2025realcam} (single-camera real-world videos) and MultiCam~\cite{bai2025recammaster} (synthetic multi-camera scenes) for stages 1 \& 2, covering both extrinsic and intrinsic control. Stage 3 uses EXIF-paired image datasets: RealBokeh~\cite{seizinger2025bokehlicious} (bokeh) and NeuralCam~\cite{ouyang2021neural} (exposure).

We build on Wan-2.1 (1.3B)~\cite{Wang2025WanOA} at 480$\times$832 resolution. We use Adam (lr $=10^{-5}$), 50 denoising steps, and train on a single RTX A6000 (96GB) for 60K/15K/6K iterations across the three stages.
\vspace{-0.3cm}

\subsection{Evaluation}
\paragraph{Baseline models.} We compare our camera-controlled VDM and provide quantitative metrics against prior works such as VACE-1.3B \cite{jiang2025vace} and CogVideoX-5B-V2V \cite{yang2024cogvideox} both of which support text prompted video-to-video generation based on a diffusion transformer backbone.

\begin{table}[b]
    \centering
    \caption{Quantitative comparison across different camera control effects. We evaluate across three categories of metrics:
  reference metrics such as PSNR and LPIPS, general video quality metrics such as VBench, and our temporal similarity metric
  wCLIP-5.}
    \label{tab:effect_eval}
    \setlength{\tabcolsep}{3pt}
    \resizebox{\columnwidth}{!}{%
    \begin{tabular}{l|l|ccc|cc|c}
    \toprule
    \multicolumn{2}{c|}{\textbf{Effect / Method}}
    & \multicolumn{3}{c|}{\textbf{Reference Metrics}}
    & \multicolumn{2}{c|}{\textbf{VBench}}
    & \textbf{Sim.} \\
    \cline{3-8}

    \multicolumn{2}{c|}{}
    & \textbf{PSNR} $\uparrow$ & \textbf{SSIM} $\uparrow$ & \textbf{LPIPS} $\downarrow$
    & \textbf{Temp.} $\uparrow$ & \textbf{Smooth} $\uparrow$
    & \textbf{wCLIP-5} $\uparrow$ \\

    \midrule\midrule

    \multirow{3}{*}{\textbf{Bokeh}}
    & VACE      & 19.86 & 0.74 & 0.36 & \textbf{0.97} & 0.98 & 0.89 \\
    & CogVideoX & 21.27 & 0.68 & 0.32 & 0.96 & \textbf{0.99} & 0.92 \\
    & \textbf{Ours} & \textbf{25.21} & \textbf{0.79} & \textbf{0.29} & 0.97 & 0.98 & \textbf{0.94} \\

    \midrule

    \multirow{3}{*}{\textbf{Motion Blur}}
    & VACE      & 20.49 & 0.73 & 0.41 & \textbf{0.97} & 0.98 & 0.89 \\
    & CogVideoX & \textbf{21.70} & 0.72 & 0.33 & 0.96 & 0.99 & 0.92 \\
    & \textbf{Ours} & 20.82 & \textbf{0.74} & \textbf{0.31} & 0.97 & \textbf{0.99} & \textbf{0.94} \\

    \midrule

    \multirow{3}{*}{\textbf{Lens Dist.}}
    & VACE      & 11.55 & 0.37 & 0.69 & 0.97 & 0.98 & 0.83 \\
    & CogVideoX & 12.27 & 0.40 & 0.62 & 0.96 & \textbf{0.98} & 0.88 \\
    & \textbf{Ours} & \textbf{13.92} & \textbf{0.44} & \textbf{0.49} & \textbf{0.98} & 0.98 & \textbf{0.91} \\

    \midrule

    \multirow{3}{*}{\textbf{Exp. Time}}
    & VACE      & 15.35 & \textbf{0.75} & 0.30 & \textbf{0.98} & 0.98 & 0.90 \\
    & CogVideoX & 16.14 & 0.65 & 0.29 & 0.96 & \textbf{0.99} & 0.94 \\
    & \textbf{Ours} & \textbf{21.08} & 0.74 & \textbf{0.22} & 0.97 & 0.98 & \textbf{0.96} \\

    \midrule

    \multirow{3}{*}{\textbf{Color Temp.}}
    & VACE      & 12.42 & 0.55 & 0.50 & \textbf{0.98} & \textbf{0.99} & 0.84 \\
    & CogVideoX & 19.30 & 0.71 & 0.26 & 0.97 & \textbf{0.99} & 0.95 \\
    & \textbf{Ours} & \textbf{23.51} & \textbf{0.90} & \textbf{0.11} & 0.98 & 0.98 & \textbf{0.98} \\

    \midrule

    \multirow{3}{*}{\textbf{Focal Len.}}
    & VACE      & 11.14 & 0.40 & 0.67 & \textbf{0.97} & 0.98 & 0.84 \\
    & CogVideoX & 11.61 & 0.43 & 0.62 & 0.96 & \textbf{0.99} & 0.87 \\
    & \textbf{Ours} & \textbf{16.51} & \textbf{0.55} & \textbf{0.33} & 0.96 & 0.98 & \textbf{0.96} \\

    \midrule\midrule

    \multirow{3}{*}{\textbf{Average}}
    & VACE      & 15.13 & 0.59 & 0.49 & \textbf{0.97} & 0.98 & 0.87 \\
    & CogVideoX & 17.05 & 0.60 & 0.41 & 0.96 & \textbf{0.99} & 0.91 \\
    & \textbf{Ours} & \textbf{20.18} & \textbf{0.69} & \textbf{0.29} & 0.97 & 0.98 & \textbf{0.95} \\


    \bottomrule
    \end{tabular}%
    }
  \end{table}

\paragraph{Metrics.} We report reference-based fidelity (PSNR, SSIM, LPIPS) against synthetic ground-truth videos, and general video quality and temporal smoothness using VBench~\cite{Huang2023VBenchCB}. Because these metrics evaluate high-level features and lack sensitivity to fine-grained optical effects, we introduce a targeted metric, wCLIP-5.

\paragraph{wCLIP-5 (Windowed CLIP Similarity).} To accommodate minor temporal jitter without allowing semantic divergence, we compute a windowed cosine similarity over a $\pm 2$ frame neighborhood. Let $\mathcal{E}_{\text{CLIP}}(\cdot)$ denote the CLIP image encoder; given generated frames $\{\hat{\mathbf{x}}_t\}_{t=1}^{T}$ and reference frames $\{\mathbf{x}_t\}_{t=1}^{T}$,
\begin{equation}
    \text{wCLIP-5} = \frac{1}{T} \sum_{t=1}^{T} \max_{k \in \{-2,\dots,2\}} \cos\!\left(\mathcal{E}_{\text{CLIP}}(\hat{\mathbf{x}}_t), \mathcal{E}_{\text{CLIP}}(\mathbf{x}_{t+k})\right),
    \vspace{-0.2cm}
\end{equation}
where out-of-bounds indices are ignored. The $\pm 2$ window relaxes strict temporal alignment, so gains on this metric reflect semantic rather than phase alignment.

  \begin{table}[t]
    \centering
    \caption{\textit{Per-effect style extraction performance.}
  NCC measures trajectory prediction accuracy (higher is better);
  Style InfoNCE measures scene independence (lower is better; random ceiling $\approx 4.07$);
  Valid fraction is the proportion of evaluation samples in which the ground-truth
  camera parameter varies over time (GT std $> 10^{-3}$), i.e., the fraction for
  which NCC is defined during training. Baselines use c-RADIO v4:
  B1: NCC-only loss, w/o temporal encoder.
  B2: NCC-only loss, w/ temporal encoder.
  }
    \label{tab:style_extraction}
    \footnotesize
    \setlength{\tabcolsep}{2pt}
    \begin{tabular}{llccccc}
    \toprule
    \textbf{Effect} & \textbf{Param.} &
    \makecell{\textbf{B1} \\ \textbf{NCC} $\uparrow$} &
    \makecell{\textbf{B2} \\ \textbf{NCC} $\uparrow$} &
    \makecell{\textbf{Ours} \\ \textbf{NCC} $\uparrow$} &
    \makecell{\textbf{Style (ours)} \\ \textbf{InfoNCE} $\downarrow$} &
    \makecell{\textbf{Valid} \\ \textbf{Frac.} $\uparrow$} \\
    \midrule
    Focal length    & F           & 0.32 & 0.75 & \textbf{0.84} & 0.38 & 0.90 \\
    Lens distortion & $\xi$       & 0.22 & 0.25 & \textbf{0.45} & 0.47 & 1.00 \\
    \addlinespace[0.5em]
    \multirow{2}{*}{Bokeh}
      & Aperture    & 0.27 & 0.35 & \textbf{0.82} & \multirow{2}{*}{2.84} & 0.76 \\
      & Focus dist. & 0.13 & 0.20 & \textbf{0.43} &                       & 0.82 \\
    \addlinespace[0.5em]
    Exposure        & EV          & 0.77 & 0.80 & \textbf{0.89} & 0.03 & 0.80 \\
    Motion blur     & Shutter     & 0.09 & 0.25 & \textbf{0.61} & 2.93 & 0.93 \\
    Color temp.     & Color temp. & 0.78 & 0.83 & \textbf{0.90} & 0.03 & 0.79 \\
    \bottomrule
    \end{tabular}
\end{table}

\subsection{Results} 
\label{sec:results}

We systematically evaluate DeltaCam's ability to render precise camera effects, mix multiple parameters, and extract reusable styles from reference footage.

\vspace{-0.2cm}
\subsubsection{Single-Effect Camera Control}
\begin{figure*}[!h]
\includegraphics[width=\textwidth]{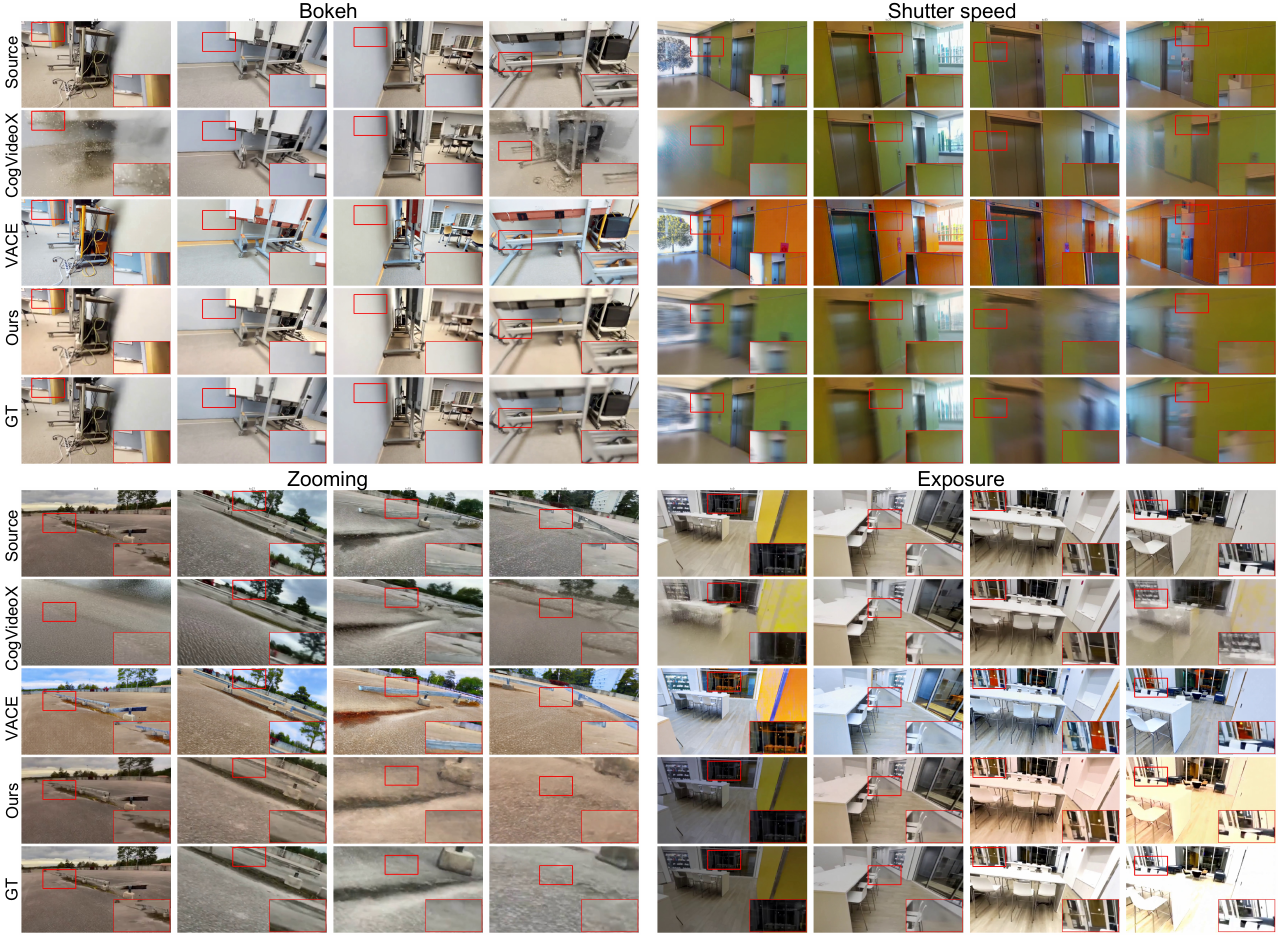}
    \caption{\textit{Visual Comparisons}. We evaluate our camera controlled generation model on several complex photographic arcs against prior state-of-the-art works.}
    \label{fig:maincomp}
    \vspace{-0.2cm}
\end{figure*}

We first evaluate the precision of DeltaCam in rendering isolated photographic effects against state-of-the-art text-to-video baselines, VACE and CogVideoX. As qualitatively shown in \cref{fig:maincomp}, our method accurately matches the synthetic ground truth trajectories for effects such as bokeh, exposure, and zooming. In contrast, text-prompted baselines struggle to render these physical transitions smoothly, often ignoring the prompt or hallucinating scene changes. 

Quantitatively (\cref{tab:effect_eval}), DeltaCam achieves the strongest fidelity on average---improving PSNR by 3.1\,dB and LPIPS by 0.12 over the next-best baseline, while maintaining comparable VBench temporal scores. Per-effect, DeltaCam leads on five of six categories across PSNR, SSIM, and LPIPS, with the largest margins on photometric effects (Color Temp.\ +4.2\,dB, Exp.\ Time +5.0\,dB). For \emph{geometric} effects (lens distortion, focal length, motion blur), reference-based metrics like PSNR and LPIPS are inherently less discriminative: small spatial displacements between rendered and ground-truth frames are penalized at full strength even when the underlying effect is faithfully reproduced. The wCLIP-5 metric, which operates on semantic features within a temporal window, better isolates effect fidelity in these cases and shows DeltaCam consistently leading (e.g., 0.91 vs.\ 0.88 on lens distortion; 0.96 vs.\ 0.87 on focal length). Matched VBench scores further confirm that injecting explicit $\Delta$-parameterized control preserves the base model's inherent temporal consistency and spatial smoothness.
\vspace{-0.15cm}

\subsubsection{Full Camera Control and Style Mixing}
\begin{figure}[!t]
\includegraphics[width=0.95\columnwidth]{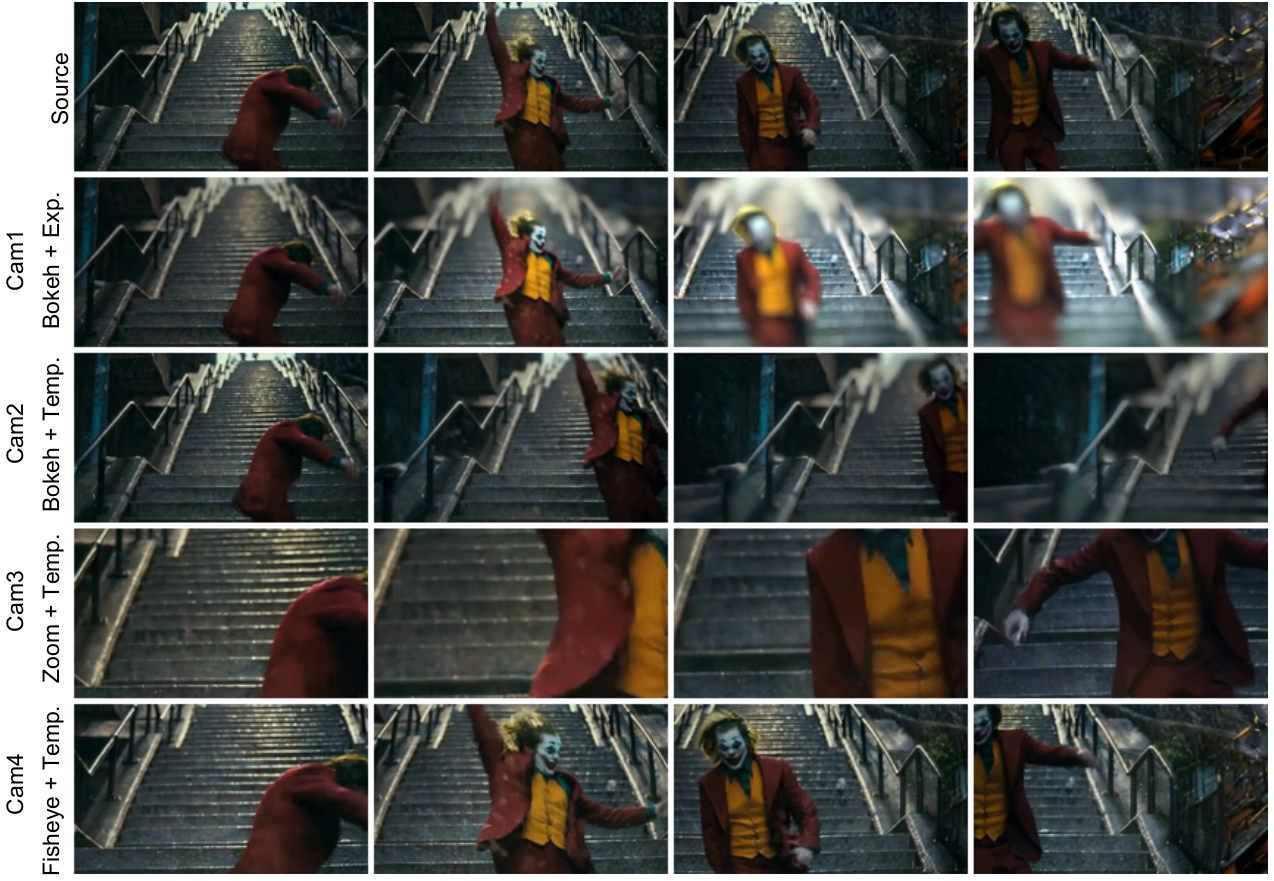}
    \caption{\textit{Full camera control}. By separating the intrinsic conditioning from extrinsic conditioning we achieve novel view generation by specifying novel camera pose trajectories while also simultaneously controlling photographic concepts as well.}
    \label{fig:camfollow}
\end{figure}

\begin{figure}[h]
\includegraphics[width=0.95\columnwidth]{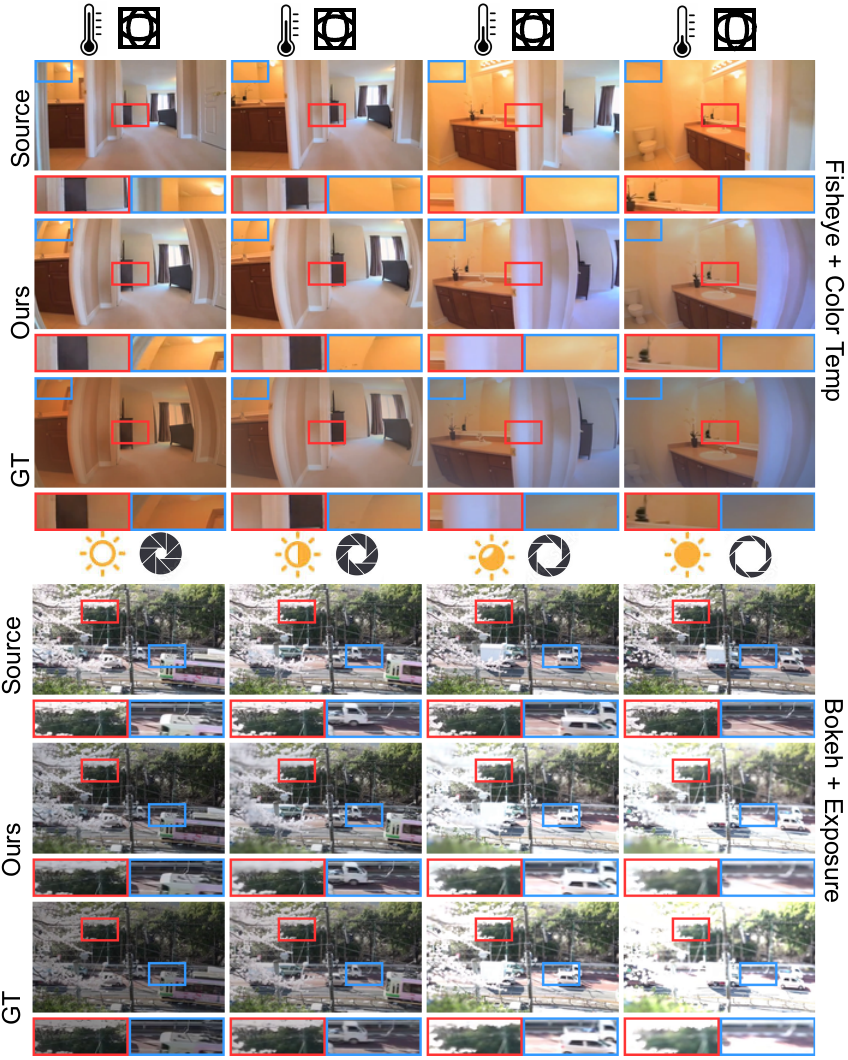}
    \caption{\textit{Style mixing}. Joint camera control of several photographic effects as a result of the latent disentanglement process employed during model training.}
    \label{fig:stylemix}
    \vspace{-0.2cm}
\end{figure}

Beyond isolated parameters, real-world cinematography requires the joint manipulation of multiple variables. In \cref{fig:camfollow}, we demonstrate full camera control by applying four distinct extrinsic camera trajectories, each coupled with a different combination of intrinsic effects, to a single source video. DeltaCam successfully disentangles these signals, generating temporally consistent novel views while simultaneously applying the requested optical shifts. We further push this capability in \cref{fig:stylemix}, demonstrating complex style mixing where two distinct photographic parameters (e.g., focal length and color temperature, or bokeh and exposure) are smoothly modulated at the same time without mutual interference or degradation of the underlying scene dynamics.

\subsubsection{Camera Style Extraction}
\begin{figure*}[h]
    \includegraphics[width=0.91\linewidth]{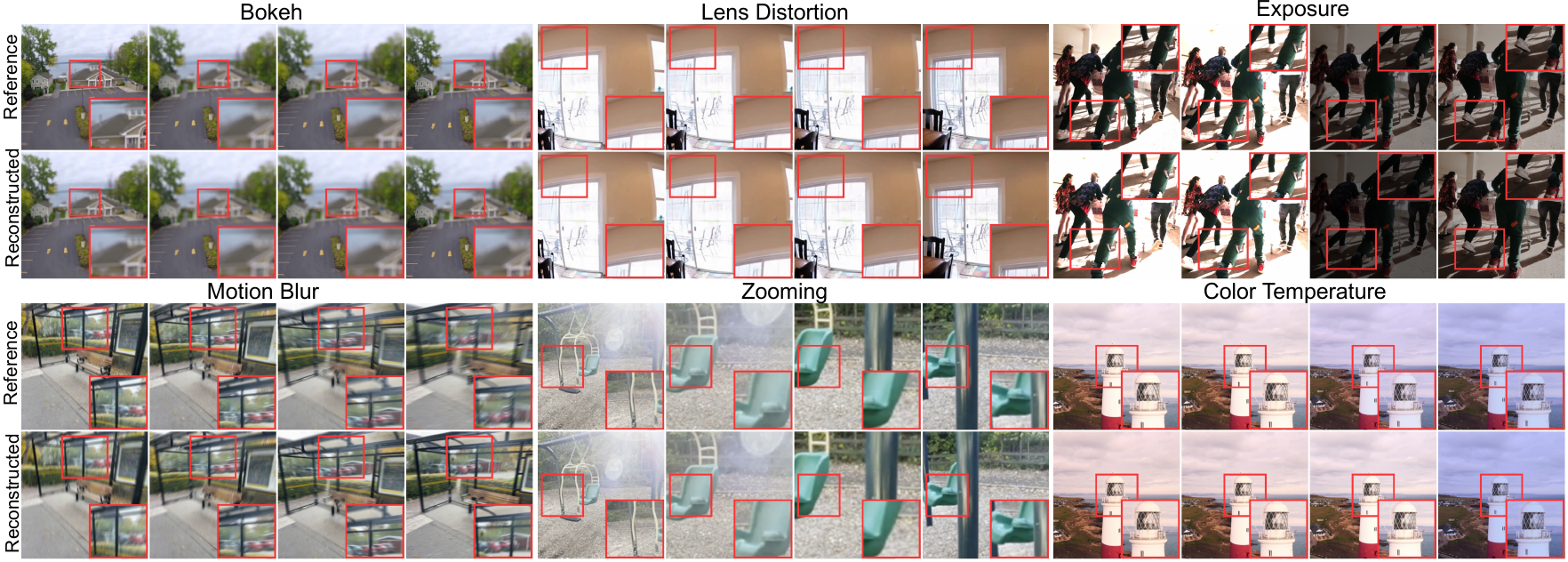}
    \caption{\textit{Camera style extraction.} We show styles extracted from video frames across timestep with photographic styles re-applied on the source video to check consistency. Our model consistently follows the style trajectory in the reference video across multiple timestamps for every effect.}
    \label{fig:fingerprint}
    \vspace{-0.2cm}
\end{figure*}

To evaluate reference-driven capabilities, we test our model's ability to invert camera effects from existing footage without explicit metadata. \cref{fig:fingerprint} visualizes the accuracy of our style extraction module: we extract the intrinsic trajectory from a reference video and re-apply it to its original source frame. As highlighted by the detailed insets, DeltaCam faithfully reconstructs the temporal evolution of the reference's optical style.


Quantitatively (\cref{tab:style_extraction}), our network achieves high Normalized Cross-Correlation (NCC) for photometric effects like color temperature (0.90) and exposure (0.89), with near-perfect scene disentanglement (Style InfoNCE $\approx$ 0.03). We compare against two c-RADIO~v4 baselines that use only the NCC trajectory loss: B1 (per-frame) and B2 (with our temporal encoder). Our full method outperforms both across every effect, with the largest gains on geometric and spatially coupled effects (focal length 0.32/0.75$\to$0.84, aperture 0.27/0.35$\to$0.82, motion blur 0.09/0.25$\to$0.61). The small B1$\to$B2 gap relative to the larger B2$\to$Ours gap confirms that our disentanglement objectives---not architecture alone---isolate the camera signal from scene content.

For spatially dependent effects like Bokeh and motion blur, trajectory prediction remains strong (aperture NCC 0.82) but Style InfoNCE only partially decreases from the random ceiling (2.84 and 2.93 vs.\ 4.07). This is expected: defocus and motion blur are physically coupled to scene depth and object velocity, so perfect content/style separation is fundamentally limited rather than a model failure. The trajectory head nonetheless decodes the parameter curve accurately, indicating that $z_{\text{style}}$ encodes the right signal even when not fully scene-orthogonal. The practical implication for inference is that $z_{\text{style}}$ can act as a ``leaky'' embedding for spatially coupled effects: when transferring bokeh from a reference video, the embedding may carry residual appearance traits from the source scene. This leakage is generally tolerable for structural effects like aperture and focus, and is largely avoided for photometric effects in isolation; cross-group leakage is discussed in the supplementary.







\begin{figure}[t]
    \includegraphics[width=0.95\columnwidth]{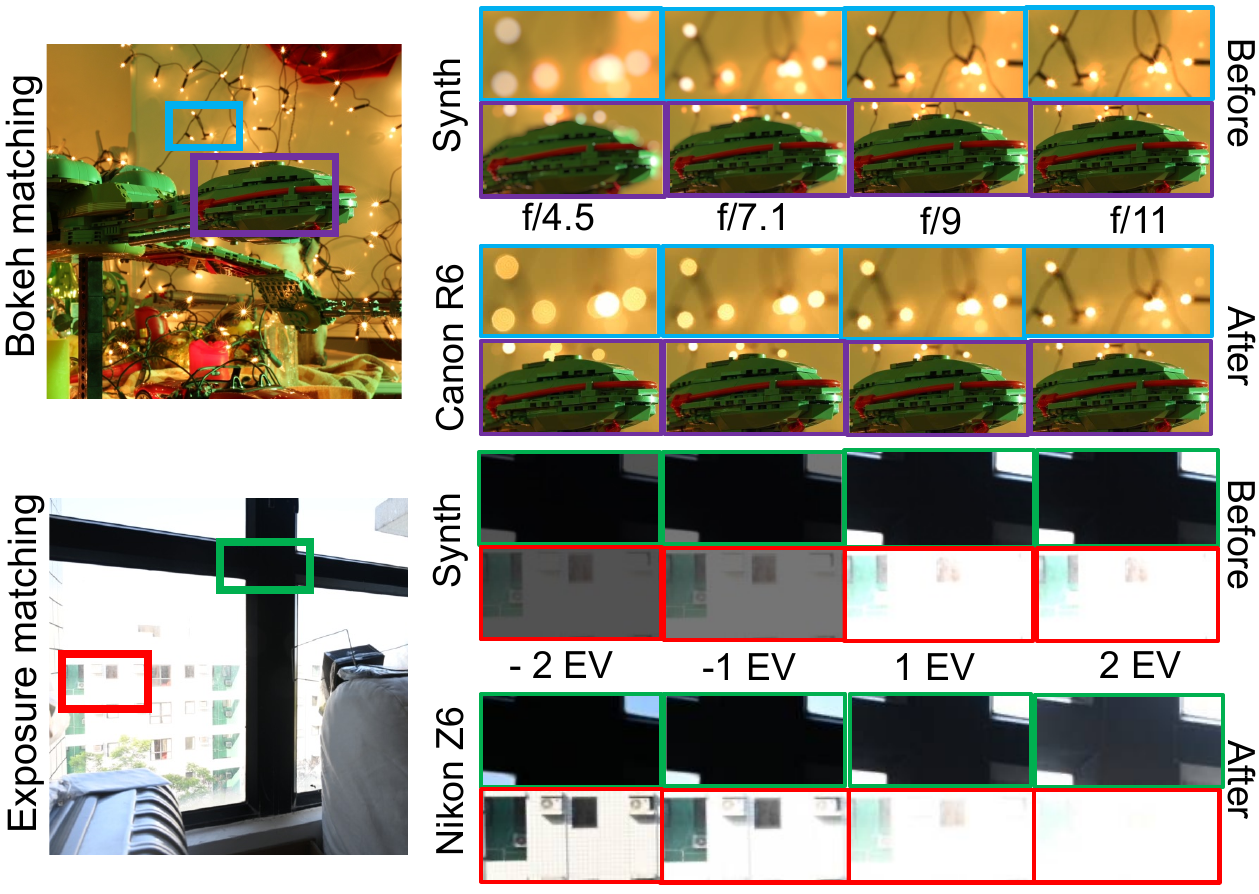}
    \caption{\textit{Real Camera Shot Matching}. Our camera matching step (Bokeh $\rightarrow$ RealBokeh; exposure $\rightarrow$ NeuralCam) matches the shot profile for camera effects from real camera paired EXIF captures making it realistic.}
    \label{fig:cam_change}
\end{figure}

\begin{figure}[h]
\includegraphics[width=0.95\columnwidth]{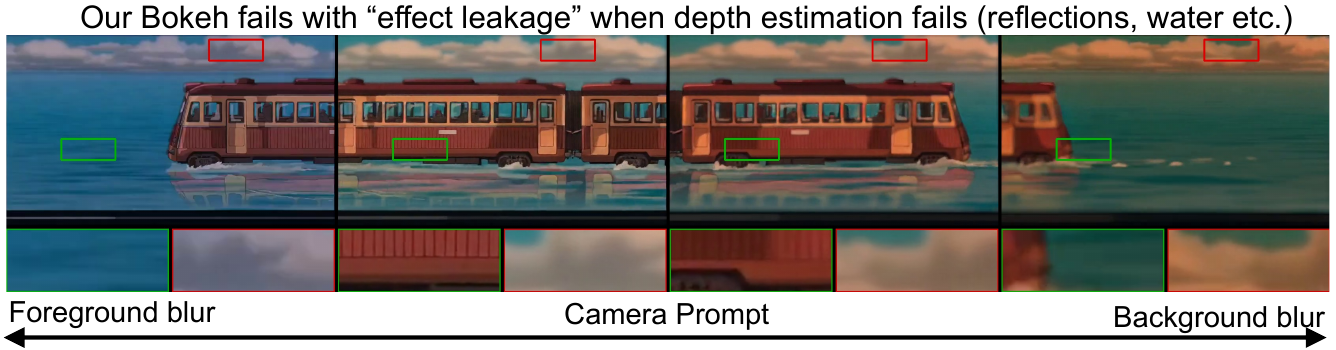}
    \caption{\textit{Limitations}. Inaccurately estimated spatial cues affects DeltaCam.}
    \label{fig:limitations}
    \vspace{-0.2cm}
\end{figure}

\begin{figure}[t]
    \includegraphics[width=\columnwidth]{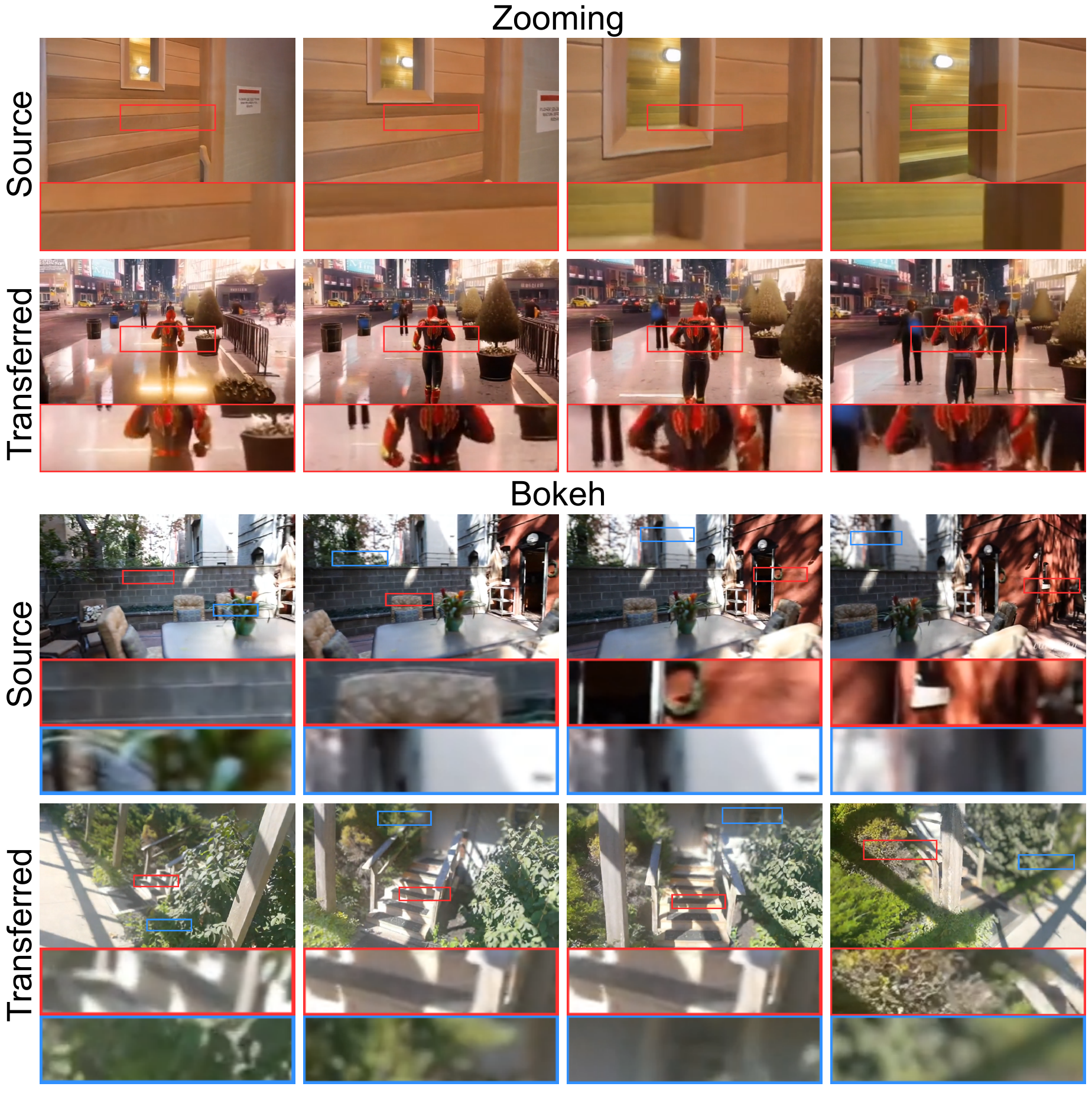}
    \caption{\textit{Camera Style transfer.} Video-to-video camera style transfer using style extraction and transfer to a new video. Zoomed insets indicate regions at similar depth or zooming direction.}
    \label{fig:style_transfer_results}
    \vspace{-0.2cm}
\end{figure}
\begin{figure}[h]
\includegraphics[width=\columnwidth]{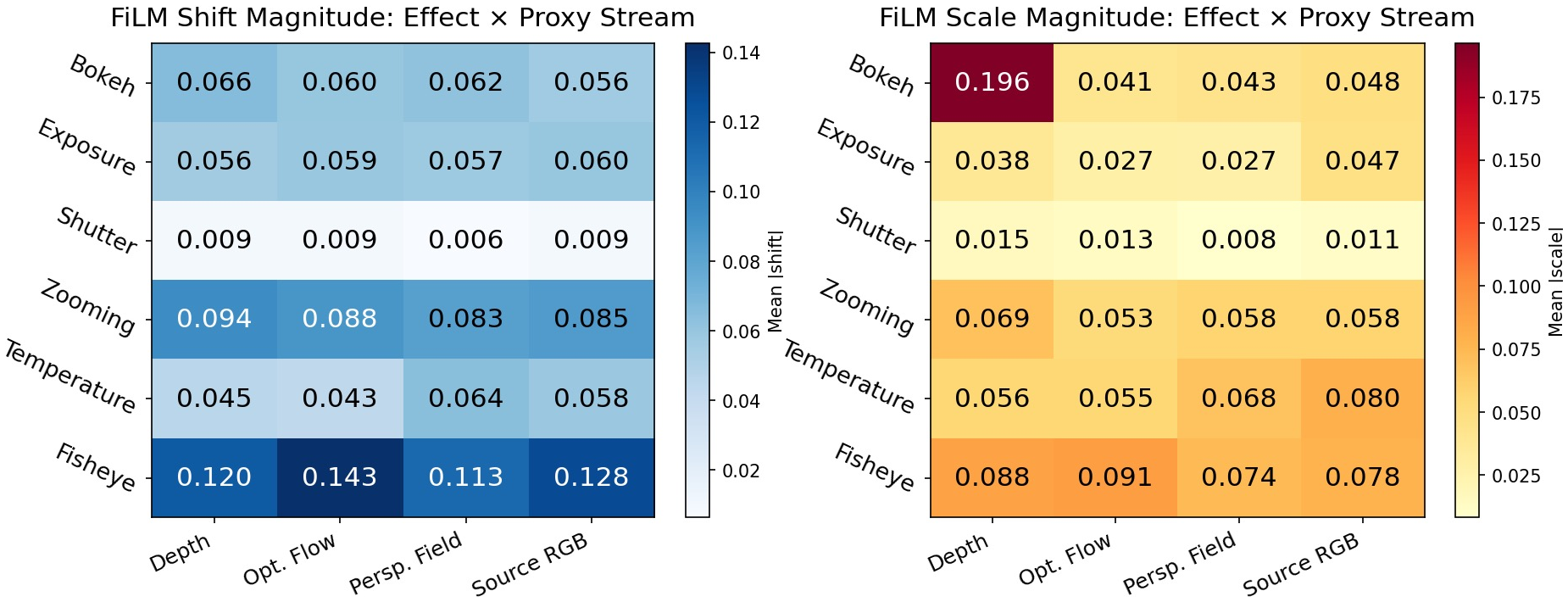}
    \caption{\textit{Ablations}. FiLM modulation strength by effect and proxy stream. Photometric effects rely on RGB; geometric effects rely on depth and perspective field.}
    \label{fig:ablations}
    \vspace{-0.4cm}
\end{figure}

\vspace{-0.1cm}
\subsubsection{Applications}
We demonstrate two practical use cases. \textbf{Real camera shot matching} (\cref{fig:cam_change}): grounding DeltaCam to specific cameras (Sony A7, Nikon Z6, Canon R6) via the EXIF tokenizer not only provides absolute parameter control but also improves perceptual quality leading to more natural Bokeh falloff and better exposure dynamic range, whereas the synthetic-trained base model shows visible artifacts. Critically, this adapts to new cameras with $<$0.1\% additional parameters, no backbone retraining. \textbf{Video-to-video style transfer} (\cref{fig:style_transfer_results}): the style extractor infers an effect trajectory from a reference video and re-applies it to a different source, transferring photographic intent (e.g., a cinematic bokeh ramp) without specifying explicit parameters. Together these span the practical spectrum from precise numerical to intuitive reference-driven control.

\vspace{-0.1cm}

\subsection{Ablation Studies}
Table~\ref{tab:ablation} reports results for three design choices.
  Zeroing individual 2D proxy streams produces minor changes ($\leq 0.15$\,dB PSNR, $\leq 0.004$ LPIPS), with source RGB causing the only measurable perceptual shift (LPIPS 0.208 vs.\ 0.204). This indicates the model treats the proxy streams as redundant geometric cues rather than relying on any single one, with RGB carrying the dominant appearance signal. Replacing the $\Delta$-parameterization with unnormalized \emph{absolute} parameters
  collapses performance dramatically
  (23.29$\rightarrow$11.88\,dB PSNR; SSIM 0.962$\rightarrow$0.360), as the model fails to map absolute camera parameters to the scene state at starting frame. Finally, simultaneous Bokeh+Exposure conditioning achieves 21.57\,dB, within $\sim$1.4\,dB of single-effect baselines (22.94 and 20.35\,dB), with the small gap absorbed on the harder single effect (Exposure). This shows the group-disentangled FiLM architecture handles multi-effect inputs with minimal interference.

\begin{table}[t]
    \centering
    \caption{Ablation study on DeltaCam. Metrics averaged over all effects.}
    \label{tab:ablation}

    \begin{subtable}[t]{\columnwidth}
    \subcaption{Component ablation performed over an ablation split separate from \cref{tab:effect_eval}.}
    \resizebox{0.9\columnwidth}{!}{%
    \begin{tabular}{lccc}
    \toprule
    \textbf{Variant} & \textbf{PSNR↑} & \textbf{SSIM↑} & \textbf{LPIPS↓} \\
    \midrule
    \multicolumn{4}{l}{\textit{Proxy Stream Contribution (stream-zeroing)}} \\
    \quad Full model (all streams)     & 22.65 & 0.875 & 0.204 \\
    \quad w/o Depth stream             & 22.50 & 0.875 & 0.201 \\
    \quad w/o Optical flow             & 22.68 & 0.876 & 0.201 \\
    \quad w/o Perspective field        & 22.63 & 0.875 & 0.202 \\
    \quad w/o Source RGB               & 22.63 & 0.873 & 0.208 \\
    \quad w/o All 2D streams           & 22.47 & 0.876 & 0.200 \\
    \midrule
    \multicolumn{4}{l}{\textit{Absolute vs Delta Parameterization}} \\
    \quad Absolute (w/o range normalization) & 11.88 & 0.360 & 0.738 \\
    \quad Delta (w/ normalization)     & 23.29 & 0.962 & 0.103 \\
    \midrule
    \multicolumn{4}{l}{\textit{Multi-Effect vs.\ Single-Effect Conditioning}} \\
    \quad Single: Bokeh only           & 22.94 & 0.859 & 0.291 \\
    \quad Single: Exposure only        & 20.35 & 0.949 & 0.136 \\
    \quad Multi: Bokeh + Exposure      & 21.57 & 0.854 & 0.274 \\
    \bottomrule
    \end{tabular}}
    \end{subtable}

    \vspace{0.5em}

    \begin{subtable}[t]{0.7\columnwidth}
    \subcaption{wCLIP temporal window sensitivity.}
    \centering
    \resizebox{\linewidth}{!}{%
    \begin{tabular}{lccc}
    \toprule
    \textbf{Method} & \textbf{wCLIP-1}$\uparrow$ & \textbf{wCLIP-5}$\uparrow$ & \textbf{wCLIP-10}$\uparrow$ \\
    \midrule
    VACE            & 0.85 & 0.87 & 0.87 \\
    CogVideoX       & 0.89 & 0.91 & 0.92 \\
    \textbf{Ours}   & \textbf{0.94} & \textbf{0.95} & \textbf{0.95} \\
    \bottomrule
    \end{tabular}}
    \end{subtable}

  \end{table}
\vspace{-0.2cm}
\subsection{Limitations}
DeltaCam’s disentanglement relies on the accuracy of deterministic spatial proxies (e.g., optical flow and monocular depth). In extreme scenarios, where these estimators fail (\cref{fig:limitations}), such as scenes with heavy motion blur, severe occlusions, or complex transparent surfaces—the conditioning module may propagate geometric artifacts into the final generation. Additionally, we restrict experiments to a 1.3B backbone (Wan-2.1) for compute reasons; scaling to larger backbones and longer video sequences is left to future work.

\vspace{-0.2cm}
\section{Conclusion}
\label{sec:conclusion}


We present DeltaCam, a video generation framework enabling explicit, temporally consistent control over intrinsic photographic effects. By leveraging a novel $\Delta$-parameterized control space and a three-stage training curriculum, our approach overcomes the severe data scarcity of real-world camera-paired videos. DeltaCam supports fine-grained optical control (e.g., bokeh, exposure ramps, dolly zooms), reference-driven style extraction, and precise real-world shot matching. Extensive evaluations demonstrate that DeltaCam significantly outperforms existing text-driven baselines in rendering accurate, disentangled camera intrinsics. Ultimately, this work establishes a scalable foundation for decoupling scene content from optical imaging behavior, advancing the capabilities of professional-grade generative cinematography.

\begin{acks}
    We are grateful to UNC Research Computing for providing access to the Longleaf cluster, on which the experiments in this work were conducted.
\end{acks}

\bibliographystyle{templates/acmart/acmart}
\bibliography{bibliography}

@article{blattmann2023stable,
  title={Stable video diffusion: Scaling latent video diffusion models to large datasets},
  author={Blattmann, Andreas and Dockhorn, Tim and Kulal, Sumith and Mendelevitch, Daniel and Kilian, Maciej and Lorenz, Dominik and Levi, Yam and English, Zion and Voleti, Vikram and Letts, Adam and others},
  journal={arXiv preprint arXiv:2311.15127},
  year={2023}
}

@article{yang2024cogvideox,
  title={Cogvideox: Text-to-video diffusion models with an expert transformer},
  author={Yang, Zhuoyi and Teng, Jiayan and Zheng, Wendi and Ding, Ming and Huang, Shiyu and Xu, Jiazheng and Yang, Yuanming and Hong, Wenyi and Zhang, Xiaohan and Feng, Guanyu and others},
  journal={arXiv preprint arXiv:2408.06072},
  year={2024}
}

@inproceedings{bahmani2025ac3d,
  title={Ac3d: Analyzing and improving 3d camera control in video diffusion transformers},
  author={Bahmani, Sherwin and Skorokhodov, Ivan and Qian, Guocheng and Siarohin, Aliaksandr and Menapace, Willi and Tagliasacchi, Andrea and Lindell, David B and Tulyakov, Sergey},
  booktitle={Proceedings of the Computer Vision and Pattern Recognition Conference},
  pages={22875--22889},
  year={2025}
}

@article{zhou2025stable,
  title={Stable virtual camera: Generative view synthesis with diffusion models},
  author={Zhou, Jensen and Gao, Hang and Voleti, Vikram and Vasishta, Aaryaman and Yao, Chun-Han and Boss, Mark and Torr, Philip and Rupprecht, Christian and Jampani, Varun},
  journal={arXiv preprint arXiv:2503.14489},
  year={2025}
}

@inproceedings{li2025realcam,
  title={Realcam-i2v: Real-world image-to-video generation with interactive complex camera control},
  author={Li, Teng and Zheng, Guangcong and Jiang, Rui and Zhan, Shuigen and Wu, Tao and Lu, Yehao and Lin, Yining and Deng, Chuanyun and Xiong, Yepan and Chen, Min and others},
  booktitle={Proceedings of the IEEE/CVF International Conference on Computer Vision},
  pages={28785--28796},
  year={2025}
}

@article{sitzmann2021light,
  title={Light field networks: Neural scene representations with single-evaluation rendering},
  author={Sitzmann, Vincent and Rezchikov, Semon and Freeman, Bill and Tenenbaum, Josh and Durand, Fredo},
  journal={Advances in Neural Information Processing Systems},
  volume={34},
  pages={19313--19325},
  year={2021}
}

@inproceedings{xing2025motioncanvas,
  title={Motioncanvas: Cinematic shot design with controllable image-to-video generation},
  author={Xing, Jinbo and Mai, Long and Ham, Cusuh and Huang, Jiahui and Mahapatra, Aniruddha and Fu, Chi-Wing and Wong, Tien-Tsin and Liu, Feng},
  booktitle={Proceedings of the Special Interest Group on Computer Graphics and Interactive Techniques Conference Conference Papers},
  pages={1--11},
  year={2025}
}

@article{yu2024viewcrafter,
  title={Viewcrafter: Taming video diffusion models for high-fidelity novel view synthesis},
  author={Yu, Wangbo and Xing, Jinbo and Yuan, Li and Hu, Wenbo and Li, Xiaoyu and Huang, Zhipeng and Gao, Xiangjun and Wong, Tien-Tsin and Shan, Ying and Tian, Yonghong},
  journal={arXiv preprint arXiv:2409.02048},
  year={2024}
}

@inproceedings{wang2025akira,
  title={AKiRa: Augmentation Kit on Rays for optical video generation},
  author={Wang, Xi and Courant, Robin and Christie, Marc and Kalogeiton, Vicky},
  booktitle={Proceedings of the Computer Vision and Pattern Recognition Conference},
  pages={2609--2619},
  year={2025}
}

@article{liao2025thinking,
  title={Thinking with Camera: A Unified Multimodal Model for Camera-Centric Understanding and Generation},
  author={Liao, Kang and Wu, Size and Wu, Zhonghua and Jin, Linyi and Wang, Chao and Wang, Yikai and Wang, Fei and Li, Wei and Loy, Chen Change},
  journal={arXiv preprint arXiv:2510.08673},
  year={2025}
}

@inproceedings{fang2024camera,
  title={Camera settings as tokens: Modeling photography on latent diffusion models},
  author={Fang, I-Sheng and Han, Yue-Hua and Chen, Jun-Cheng},
  booktitle={SIGGRAPH Asia 2024 Conference Papers},
  pages={1--11},
  year={2024}
}

@article{kim2025ccmnet,
  title={CCMNet: Leveraging Calibrated Color Correction Matrices for Cross-Camera Color Constancy},
  author={Kim, Dongyoung and Afifi, Mahmoud and Kim, Dongyun and Brown, Michael S and Kim, Seon Joo},
  journal={arXiv preprint arXiv:2504.07959},
  year={2025}
}

@inproceedings{guo2025depth,
  title={Depth any camera: Zero-shot metric depth estimation from any camera},
  author={Guo, Yuliang and Garg, Sparsh and Miangoleh, S Mahdi H and Huang, Xinyu and Ren, Liu},
  booktitle={Proceedings of the Computer Vision and Pattern Recognition Conference},
  pages={26996--27006},
  year={2025}
}

@inproceedings{ouyang2021neural,
  title={Neural camera simulators},
  author={Ouyang, Hao and Shi, Zifan and Lei, Chenyang and Law, Ka Lung and Chen, Qifeng},
  booktitle={Proceedings of the IEEE/CVF conference on computer vision and pattern recognition},
  pages={7700--7709},
  year={2021}
}

@inproceedings{yuan2025generative,
  title={Generative photography: Scene-consistent camera control for realistic text-to-image synthesis},
  author={Yuan, Yu and Wang, Xijun and Sheng, Yichen and Chennuri, Prateek and Zhang, Xingguang and Chan, Stanley},
  booktitle={Proceedings of the Computer Vision and Pattern Recognition Conference},
  pages={7920--7930},
  year={2025}
}

@inproceedings{bernal2025precisecam,
  title={PreciseCam: Precise Camera Control for Text-to-Image Generation},
  author={Bernal-Berdun, Edurne and Serrano, Ana and Masia, Belen and Gadelha, Matheus and Hold-Geoffroy, Yannick and Sun, Xin and Gutierrez, Diego},
  booktitle={Proceedings of the Computer Vision and Pattern Recognition Conference},
  pages={2724--2733},
  year={2025}
}

@inproceedings{zhang2023adding,
  title={Adding conditional control to text-to-image diffusion models},
  author={Zhang, Lvmin and Rao, Anyi and Agrawala, Maneesh},
  booktitle={Proceedings of the IEEE/CVF international conference on computer vision},
  pages={3836--3847},
  year={2023}
}

@inproceedings{radford2021learning,
  title={Learning transferable visual models from natural language supervision},
  author={Radford, Alec and Kim, Jong Wook and Hallacy, Chris and Ramesh, Aditya and Goh, Gabriel and Agarwal, Sandhini and Sastry, Girish and Askell, Amanda and Mishkin, Pamela and Clark, Jack and others},
  booktitle={International conference on machine learning},
  pages={8748--8763},
  year={2021},
  organization={PmLR}
}

@article{oquab2023dinov2,
  title={Dinov2: Learning robust visual features without supervision},
  author={Oquab, Maxime and Darcet, Timoth{\'e}e and Moutakanni, Th{\'e}o and Vo, Huy and Szafraniec, Marc and Khalidov, Vasil and Fernandez, Pierre and Haziza, Daniel and Massa, Francisco and El-Nouby, Alaaeldin and others},
  journal={arXiv preprint arXiv:2304.07193},
  year={2023}
}

@article{hou2024training,
  title={Training-free camera control for video generation},
  author={Hou, Chen and Chen, Zhibo},
  journal={arXiv preprint arXiv:2406.10126},
  year={2024}
}

@inproceedings{zhou2025latent,
  title={Latent-Reframe: Enabling Camera Control for Video Diffusion Models without Training},
  author={Zhou, Zhenghong and An, Jie and Luo, Jiebo},
  booktitle={Proceedings of the IEEE/CVF International Conference on Computer Vision},
  pages={12779--12789},
  year={2025}
}

@article{guhan2025cammimic,
  title={CamMimic: Zero-Shot Image To Camera Motion Personalized Video Generation Using Diffusion Models},
  author={Guhan, Pooja and Kothandaraman, Divya and Huang, Tsung-Wei and Su, Guan-Ming and Manocha, Dinesh},
  journal={arXiv preprint arXiv:2504.09472},
  year={2025}
}

@article{bai2025recammaster,
  title={Recammaster: Camera-controlled generative rendering from a single video},
  author={Bai, Jianhong and Xia, Menghan and Fu, Xiao and Wang, Xintao and Mu, Lianrui and Cao, Jinwen and Liu, Zuozhu and Hu, Haoji and Bai, Xiang and Wan, Pengfei and others},
  journal={arXiv preprint arXiv:2503.11647},
  year={2025}
}

@article{meng2021sdedit,
  title={Sdedit: Guided image synthesis and editing with stochastic differential equations},
  author={Meng, Chenlin and He, Yutong and Song, Yang and Song, Jiaming and Wu, Jiajun and Zhu, Jun-Yan and Ermon, Stefano},
  journal={arXiv preprint arXiv:2108.01073},
  year={2021}
}

@article{song2020denoising,
  title={Denoising diffusion implicit models},
  author={Song, Jiaming and Meng, Chenlin and Ermon, Stefano},
  journal={arXiv preprint arXiv:2010.02502},
  year={2020}
}

@inproceedings{gu2024videoswap,
  title={Videoswap: Customized video subject swapping with interactive semantic point correspondence},
  author={Gu, Yuchao and Zhou, Yipin and Wu, Bichen and Yu, Licheng and Liu, Jia-Wei and Zhao, Rui and Wu, Jay Zhangjie and Zhang, David Junhao and Shou, Mike Zheng and Tang, Kevin},
  booktitle={Proceedings of the IEEE/CVF Conference on Computer Vision and Pattern Recognition},
  pages={7621--7630},
  year={2024}
}

@inproceedings{wei2024dreamvideo,
  title={Dreamvideo: Composing your dream videos with customized subject and motion},
  author={Wei, Yujie and Zhang, Shiwei and Qing, Zhiwu and Yuan, Hangjie and Liu, Zhiheng and Liu, Yu and Zhang, Yingya and Zhou, Jingren and Shan, Hongming},
  booktitle={Proceedings of the IEEE/CVF Conference on Computer Vision and Pattern Recognition},
  pages={6537--6549},
  year={2024}
}

@inproceedings{gu2025diffusion,
  title={Diffusion as shader: 3d-aware video diffusion for versatile video generation control},
  author={Gu, Zekai and Yan, Rui and Lu, Jiahao and Li, Peng and Dou, Zhiyang and Si, Chenyang and Dong, Zhen and Liu, Qifeng and Lin, Cheng and Liu, Ziwei and others},
  booktitle={Proceedings of the Special Interest Group on Computer Graphics and Interactive Techniques Conference Conference Papers},
  pages={1--12},
  year={2025}
}

@inproceedings{resslerdismo,
  title={DisMo: Disentangled Motion Representations for Open-World Motion Transfer},
  author={Ressler-Antal, Thomas and Fundel, Frank and Alaya, Malek Ben and Baumann, Stefan Andreas and Krause, Felix and Gui, Ming and Ommer, Bj{\"o}rn},
  booktitle={The Thirty-ninth Annual Conference on Neural Information Processing Systems},
  year={2025}
}

@article{HaCohen2024LTXVideoRV,
  title={LTX-Video: Realtime Video Latent Diffusion},
  author={Yoav HaCohen and Nisan Chiprut and Benny Brazowski and Daniel Shalem and David-Pur Moshe and Eitan Richardson and E. I. Levin and Guy Shiran and Nir Zabari and Ori Gordon and Poriya Panet and Sapir Weissbuch and Victor Kulikov and Yaki Bitterman and Zeev Melumian and Ofir Bibi},
  journal={ArXiv},
  year={2024},
  volume={abs/2501.00103},
  url={https://api.semanticscholar.org/CorpusID:275212083}
}

@inproceedings{Perez2017FiLMVR,
  title={FiLM: Visual Reasoning with a General Conditioning Layer},
  author={Ethan Perez and Florian Strub and Harm de Vries and Vincent Dumoulin and Aaron C. Courville},
  booktitle={AAAI Conference on Artificial Intelligence},
  year={2017},
  url={https://api.semanticscholar.org/CorpusID:19119291}
}

@article{Geng2024MotionPC,
  title={Motion Prompting: Controlling Video Generation with Motion Trajectories},
  author={Daniel Geng and Charles Herrmann and Junhwa Hur and Forrester Cole and Serena Zhang and Tobias Pfaff and Tatiana Lopez-Guevara and Carl Doersch and Yusuf Aytar and Michael Rubinstein and Chen Sun and Oliver Wang and Andrew Owens and Deqing Sun},
  journal={2025 IEEE/CVF Conference on Computer Vision and Pattern Recognition (CVPR)},
  year={2024},
  pages={1-12},
  url={https://api.semanticscholar.org/CorpusID:274446282}
}

@article{Wang2025CineMasterA3,
  title={CineMaster: A 3D-Aware and Controllable Framework for Cinematic Text-to-Video Generation},
  author={Qinghe Wang and Yawen Luo and Xiaoyu Shi and Xu Jia and Huchuan Lu and Tianfan Xue and Xintao Wang and Pengfei Wan and Di Zhang and Kun Gai},
  journal={Proceedings of the Special Interest Group on Computer Graphics and Interactive Techniques Conference Conference Papers},
  year={2025},
  url={https://api.semanticscholar.org/CorpusID:276287673}
}

@article{zhang2023controlvideo,
  title={ControlVideo: Training-free Controllable Text-to-Video Generation},
  author={Zhang, Yabo and Wei, Yuxiang and Jiang, Dongsheng and Zhang, Xiaopeng and Zuo, Wangmeng and Tian, Qi},
  journal={arXiv preprint arXiv:2305.13077},
  year={2023}
}

@article{Chen2023MotionConditionedDM,
  title={Motion-Conditioned Diffusion Model for Controllable Video Synthesis},
  author={Tsai-Shien Chen and Chieh Hubert Lin and Hung-Yu Tseng and Tsung-Yi Lin and Ming Yang},
  journal={ArXiv},
  year={2023},
  volume={abs/2304.14404},
  url={https://api.semanticscholar.org/CorpusID:258352582}
}

@article{Wang2025WanOA,
  title={Wan: Open and Advanced Large-Scale Video Generative Models},
  author={Ang Wang and Baole Ai and Bin Wen and Chaojie Mao and Chen-Wei Xie and Di Chen and Feiwu Yu and Haiming Zhao and Jianxiao Yang and Jianyuan Zeng and Jiayu Wang and Jingfeng Zhang and Jingren Zhou and Jinkai Wang and Jixuan Chen and Kai Zhu and Kang Zhao and Keyu Yan and Lianghua Huang and Xiaofeng Meng and Ningying Zhang and Pandeng Li and Ping Wu and Ruihang Chu and Rui Feng and Shiwei Zhang and Siyang Sun and Tao Fang and Tianxing Wang and Tianyi Gui and Tingyu Weng and Tong Shen and Wei Lin and Wei Wang and Wei Wang and Wen-Chao Zhou and Wente Wang and Wen Shen and Wenyuan Yu and Xianzhong Shi and Xiaomin Huang and Xin Xu and Yan Kou and Yan-Mei Lv and Yifei Li and Yijing Liu and Yiming Wang and Yingya Zhang and Yitong Huang and Yong Li and You Wu and Yu Liu and Yulin Pan and Yun Zheng and Yuntao Hong and Yupeng Shi and Yutong Feng and Zeyinzi Jiang and Zhengbin Han and Zhigang Wu and Ziyu Liu},
  journal={ArXiv},
  year={2025},
  volume={abs/2503.20314},
  url={https://api.semanticscholar.org/CorpusID:277321639}
}

@article{Huang2023VBenchCB,
  title={VBench: Comprehensive Benchmark Suite for Video Generative Models},
  author={Ziqi Huang and Yinan He and Jiashuo Yu and Fan Zhang and Chenyang Si and Yuming Jiang and Yuanhan Zhang and Tianxing Wu and Qingyang Jin and Nattapol Chanpaisit and Yaohui Wang and Xinyuan Chen and Limin Wang and Dahua Lin and Yu Qiao and Ziwei Liu},
  journal={2024 IEEE/CVF Conference on Computer Vision and Pattern Recognition (CVPR)},
  year={2023},
  pages={21807-21818},
  url={https://api.semanticscholar.org/CorpusID:265506207}
}

@article{xiao2025captain,
  title={Captain cinema: Towards short movie generation},
  author={Xiao, Junfei and Yang, Ceyuan and Zhang, Lvmin and Cai, Shengqu and Zhao, Yang and Guo, Yuwei and Wetzstein, Gordon and Agrawala, Maneesh and Yuille, Alan and Jiang, Lu},
  journal={arXiv preprint arXiv:2507.18634},
  year={2025}
}

@article{tedla2025generating,
  title={Generating the Past, Present and Future from a Motion-Blurred Image},
  author={Tedla, SaiKiran and Zhu, Kelly and Canham, Trevor and Taubner, Felix and Brown, Michael S and Kutulakos, Kiriakos N and Lindell, David B},
  journal={ACM Transactions on Graphics (TOG)},
  volume={44},
  number={6},
  pages={1--15},
  year={2025},
  publisher={ACM New York, NY, USA}
}

@inproceedings{jiang2025vace,
  title={Vace: All-in-one video creation and editing},
  author={Jiang, Zeyinzi and Han, Zhen and Mao, Chaojie and Zhang, Jingfeng and Pan, Yulin and Liu, Yu},
  booktitle={Proceedings of the IEEE/CVF International Conference on Computer Vision},
  pages={17191--17202},
  year={2025}
}

@misc{ranzinger2026cradiov4techreport,
      title={C-RADIOv4 (Tech Report)}, 
      author={Mike Ranzinger and Greg Heinrich and Collin McCarthy and Jan Kautz and Andrew Tao and Bryan Catanzaro and Pavlo Molchanov},
      year={2026},
      eprint={2601.17237},
      archivePrefix={arXiv},
      primaryClass={cs.CV},
      url={https://arxiv.org/abs/2601.17237}, 
}

@inproceedings{seizinger2025bokehlicious,
  title={Bokehlicious: Photorealistic bokeh rendering with controllable apertures},
  author={Seizinger, Tim and Vasluianu, Florin-Alexandru and Conde, Marcos V and Wu, Zongwei and Timofte, Radu},
  booktitle={Proceedings of the IEEE/CVF International Conference on Computer Vision},
  pages={8908--8917},
  year={2025}
}

\appendixpageoff
\appendixtitleoff
\renewcommand{\appendixtocname}{Supplementary material}
\begin{appendices}
\crefalias{section}{supp}
\normalsize

\end{appendices}
\end{document}